\pdfoutput=1

\documentclass[11pt]{article}

\usepackage[]{ACL2023}

\usepackage{times}
\usepackage{latexsym}

\usepackage{comment}
\usepackage{rotating}
\usepackage[utf8]{inputenc}
\usepackage{times}
\usepackage{latexsym}
\usepackage{rotating,tabularx,siunitx}

\usepackage{float}
\usepackage{graphicx}
\usepackage{subfigure}
\usepackage{caption}
\usepackage{amsmath,amssymb}
\usepackage{mathtools}
\usepackage{booktabs}
\usepackage{multirow}
\usepackage{multicol}
\usepackage{siunitx}
\usepackage{adjustbox}
\usepackage{pifont}
\usepackage{tabularx}
\usepackage{enumitem}
\usepackage{array, makecell, multirow}
\newcolumntype{P}[1]{>{\centering\arraybackslash}p{#1}}
\usepackage{siunitx}

\usepackage[T1]{fontenc}

\usepackage[utf8]{inputenc}

\usepackage{microtype}

\usepackage{inconsolata}
\usepackage{xcolor,colortbl}
\usepackage{soul}
\definecolor{headcolor}{HTML}{018161}
\definecolor{relationcolor}{HTML}{d95f02}
\definecolor{tailcolor}{HTML}{6560a3}

\newcommand{\ust}{\ensuremath{^\spadesuit}}
\newcommand{\Tencent}{\ensuremath{^\dagger}}

\title{NegotiationToM: A Benchmark for Stress-testing Machine Theory of Mind on Negotiation Surrounding}

\author{Chunkit Chan\ust \ \ \ \ Cheng Jiayang\ust \ \ \ \ Yauwai Yim\ust \ \ \ \ Zheye Deng\ust \\ 
\ \ \ \ \textbf{Wei Fan\ust}  \ \ \ \ \textbf{Haoran Li\ust} \ \ \ \ \textbf{Xin Liu\ust} \ \ \ \ \textbf{Hongming Zhang\Tencent}\\ 
\ \ \ \ \textbf{Weiqi Wang\ust} \ \ \ \  \textbf{Yangqiu Song\ust} \\
\ust The Hong Kong University of Science and Technology \\
\Tencent Tencent AI Lab, Seattle\\
\texttt{\{ckchancc, yqsong\}@cse.ust.hk} \ \ \ \ 
}

\begin{document}
\maketitle
\begin{abstract}
Large Language Models (LLMs) have sparked substantial interest and debate concerning their potential emergence of Theory of Mind (ToM) ability. Theory of mind evaluations currently focuses on testing models using machine-generated data or game settings prone to shortcuts and spurious correlations, which lacks evaluation of machine ToM ability in real-world human interaction scenarios. This poses a pressing demand to develop new real-world scenario benchmarks. We introduce NegotiationToM\footnote{The dataset is available at~\url{https://github.com/HKUST-KnowComp/NegotiationToM}}, a new benchmark designed to stress-test machine ToM in real-world negotiation surrounding covered multi-dimensional mental states (i.e., desires, beliefs, and intentions). Our benchmark builds upon the Belief-Desire-Intention (BDI) agent modeling theory and conducts the necessary empirical experiments to evaluate large language models. Our findings demonstrate that NegotiationToM is challenging for state-of-the-art LLMs, as they consistently perform significantly worse than humans, even when employing the chain-of-thought (CoT) method.

\end{abstract}

\section{Introduction}
\begin{figure}[t]
    \centering
    \includegraphics[width=0.48\textwidth]{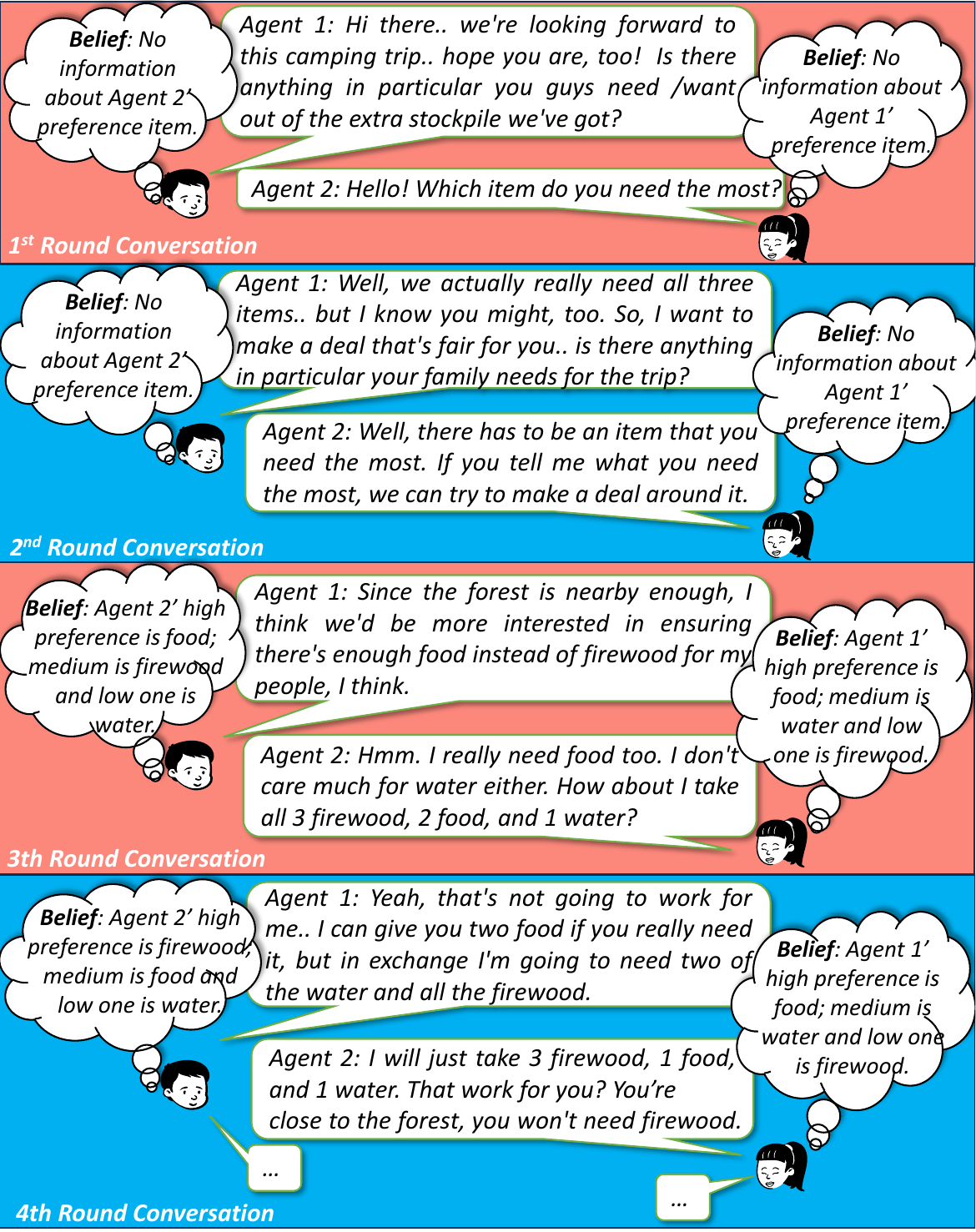}
    \vspace{-0.8cm}
    \caption{A negotiation example in NegotiationToM. Two agents are negotiating for food, water, and firewood packages for their upcoming trip.} 
    \label{fig:Example_NegotiationToM}
    \vspace{-0.7cm}
\end{figure}

Theory of Mind (ToM) was introduced as an agent’s capacity to infer the mental states of others, such as desires, beliefs, and intentions~\cite{premack1978does, DBLP:conf/emnlp/MaSPC23}. Numerous scenarios involving human cognition and social reasoning rely on the ToM modeling of others’ mental states~\cite{gopnik1992child, baron1997mindblindness,DBLP:journals/corr/abs-1810-07528}, such as comprehending and forecasting others’ actions~\cite{dennett1988precis}, planning over others’ beliefs and subsequent actions~\cite{DBLP:conf/ro-man/FavierSA23}, and various forms of reasoning and decision-making~\cite{DBLP:journals/ai/PereiraPS16,rusch2020theory}. Some previous research believes that LLMs already exhibit a high level of competence in addressing ToM tasks~\cite{strachan2024testing, DBLP:journals/corr/abs-2303-12712, DBLP:journals/corr/abs-2302-02083}, while other studies express doubt and develop benchmarks to illustrate that LLMs do not possess proficient ability in ToM tasks~\cite{DBLP:conf/emnlp/Sap0FC22, DBLP:journals/corr/abs-2302-08399, DBLP:conf/eacl/ShapiraLAZCGSS24}. However, these traditional evaluation benchmarks for language models are primarily theoretical game settings or synthetic template-based data generated by the large language model, which may inherently suffer from shortcuts and spurious correlations~\cite{DBLP:conf/acl/SclarKWS0T23, DBLP:journals/corr/abs-2302-08399, DBLP:journals/corr/abs-2305-14763, DBLP:conf/emnlp/MaSPC23}.
Consequently, these benchmarks assess language models from a theoretical perspective, which may not precisely and effectively reflect the ToM capabilities of large language models in practical situations.

In reality, ToM ability plays a crucial role in comprehending dynamic social interactions (e.g., negotiation conversations) by forming an essential element of effective communication~\cite{frith1994autism,schober2005conceptual}, and inferring other's mental states in a conversation requires machines as humans to comprehend text beyond surface forms of utterance and utilize the incomplete information presented in the conversation.
ToM is closely related to interpersonal social intelligence~\cite{ganaie2015study, stone2006theory, williams2022supporting, DBLP:conf/emnlp/Sap0FC22}, which allows us to navigate and understand social situations ranging from simple everyday interactions to complex negotiations~\cite{DBLP:conf/acl/YangCN20, DBLP:conf/emnlp/0002SZ0K0S23, DBLP:journals/aamas/WeerdVV17, gardner1995kids, DBLP:conf/prima/WeerdVV13}.

The negotiation dialogues contain complicated and diverse aspects of a realistic negotiation, such as rapport building, discussing preferences, exchanging offers, emotional expression, and persuasion with personal and logical arguments~\cite{DBLP:conf/naacl/ChawlaRCLMG21}. In a realistic negotiation, humans innately infer the mental states of the other party and proceed with their subsequent actions based on their own beliefs and desires. For example, in Figure~\ref{fig:Example_NegotiationToM}, two agents negotiate for food, water, and firewood packages for their upcoming trip. Initially, agent 1 lacks any information pertaining to the preference order of agent 2. Thus, agent 1’s belief is \textit{“no information on the agent 2 preference item”}, and agent 1 intends to elicit the item preference order from agent 2 to guide further action based on their own belief. Furthermore, belief is commonly employed to denote an individual’s cognitive stance or acceptance of something as true or holding it to be the case~\cite{turiel2008development}. This belief may undergo changes during the negotiation as perceiving more available information behind the conversation. In the fourth round of conversation depicted in Figure~\ref{fig:Example_NegotiationToM}, agent 1’s dynamic belief changed from \textit{“Agent 2’ high preference is food, medium preference is firewood, and the low one is water”} to \textit{“Agent 2’s high preference is firewood, medium preference is food, and the low one is water.”} Therefore, negotiation serves as an ideal scenario to assess the theory of mind ability of language learning models in the real world due to its complexity and the linguistic diversity inherent in negotiation conversations.

In this work, we introduce NegotiationToM, a natural conversational benchmark for stress-testing machine ToM in real-world negotiation surroundings involving multi-dimensional mental states (i.e., desires, beliefs, and intentions), inspired by the Belief-Desire-Intention (BDI) agent model proposed by~\citet{bratman1987intention}. 
The goal of NegotiationToM is to effectively measure how well large language models (LLMs) can track the mental states of negotiation participants in conversations and evaluate LLMs' capability for a coherent understanding of others’ mental states in the conversation context where there are gradually increasing rounds of utterance (i.e., increase available information). We hope our benchmark and experimental results in the real-world scenario complement the prior theoretical works, which yield important insights into the intensive debate around ToM~\cite{whang2023can} in LLMs. 
Our contributions are summarized as follows:
{\begin{itemize}[leftmargin=*]
    \item To the best of our knowledge, NegotiationToM is the first human-annotated natural conversational benchmark to introduce negotiation theory of mind evaluation for large language models in realistic negotiations.
    \item Our benchmark covered multi-dimensional mental states (i.e., desires, beliefs, and intentions) to assess how well large language models can track the mental states of negotiation participants in conversations and coherent understanding of others’ mental states with increased available and accessible information.
    \item We undertake the necessary empirical experiments to evaluate large language models (LLMs) on the NegotiationToM benchmark and conduct extensive in-depth analysis to explore the LLMs' empirical performance under various settings.
\end{itemize}}

\section{Related Work}
\paragraph{Theory of Mind Benchmarks}
The existing ToM evaluation benchmarks for large language models are primarily synthetic template-based data generated~\cite{DBLP:conf/emnlp/0002SZ0K0S23, DBLP:journals/corr/abs-2306-15448} or derived from the Sally-Anne False Belief Test~\cite{baron1985does, DBLP:conf/emnlp/NematzadehBGGG18, DBLP:conf/cogsci/GrantNG17, DBLP:conf/emnlp/LeBN19,DBLP:journals/corr/abs-2310-03051},  which assesses model ability from a theoretical perspective and may inherently suffer from shortcuts and spurious correlations~\cite{DBLP:conf/acl/SclarKWS0T23, DBLP:journals/corr/abs-2302-08399, DBLP:journals/corr/abs-2305-14763, DBLP:conf/emnlp/MaSPC23}. Other works, such as~\citet{DBLP:conf/acl/ShapiraZG23} build benchmarks based on the Faux Pas Test~\cite{baron1999recognition}. The most related work to ours is the BigToM benchmark proposed by~\cite{DBLP:journals/corr/abs-2306-15448}, which presents a framework for designing a ToM benchmark from synthetic templates for evaluating different aspects of LLMs’ ToM capabilities (e.g., desire and belief). However, this work and other theoretical benchmarks may not reflect the ToM capabilities of large language models in real-world scenarios. Moreover, most of these prior works are concentrated on the belief aspects of the Theory of Mind. Therefore, this work introduces NegotiationToM, which is a multi-category mental state benchmark in realistic negotiation scenarios. 

\paragraph{Negotiation}
Negotiation is an expanding area of research in the natural language processing field, and \citet{zhan2022let} conducted an impressive survey of existing literature on dialogue systems for negotiation. 
\citet{lewis2017deal} train recurrent neural networks to generate natural language dialogues in negotiations. \citet{he2018decoupling} proposed a modular generative model that is based on dialogue acts. 
Various disciplines have explored bilateral bargaining from diverse perspectives and employing different methodologies. Economic theory has examined the influence of incomplete information \cite{ausubel2002bargaining} and emphasized the significance of explicit communication \cite{crawford1990explicit,roth2020}.
\citet{bazerman2000negotiation} and \citet{pruitt2013negotiation} present a comprehensive overview of the psychology research on negotiation. These previous studies generally neglect the content of communication, although there are a few noteworthy exceptions \cite{swaab2011early,jeong2019communicating,lee2017can, he2018decoupling, DBLP:conf/acl/HeddayaDTVZ23}. One intriguing work by \citet{DBLP:conf/acl/YangCN20} introduces a probabilistic formulation method to encapsulate the opponent’s personality type during learning and inference, drawing inspiration from the idea of incorporating a theory of mind (ToM) into machines. However, distinct from this approach, our work presents a benchmark integrating a theory of mind (ToM) into the negotiation surroundings.


\begin{table*}[th]
\centering


\scalebox{0.65}{
\begin{tabular}{P{4cm}|P{8cm}}
\hline
\multicolumn{2}{c}{\textbf{NegotiationToM Questions}} \\ \hline
\multicolumn{1}{l|}{\textbf{Desire Question}} & \multicolumn{1}{l}{What is \textcolor{headcolor}{<Agent 1/Agent2>}’s \textcolor{tailcolor}{<high/medium/low>} preference for  items given the dialogue history?}\\ \hline
\multicolumn{1}{l|}{\textbf{Belief Question}} & \multicolumn{1}{l}{What is \textcolor{tailcolor}{<high/medium/low>} preference for items \textcolor{headcolor}{<Agent 1/Agent2>} thinks  \textcolor{headcolor}{<Agent 2/Agent1>} is given the dialogue history?}\\ \hline
\multicolumn{1}{l|}{\textbf{Intention Question}} & \multicolumn{1}{l}{What is intentions of \textcolor{headcolor}{<Agent 1/Agent2>}’s expressed in \textcolor{tailcolor}{<Utterance>} given the dialogue history?}\\ \hline
\end{tabular}}\\

\vspace{0.1cm}

\scalebox{0.6}{
\begin{tabular}{p{9cm}|p{3.2cm}|p{5.5cm}|p{5.5cm}}
\hline
\multicolumn{1}{c|}{\textbf{Conversation}} & \multicolumn{1}{c|}{\textbf{Intention}} & \multicolumn{1}{c|}{\textbf{Belief}} & \multicolumn{1}{c}{\textbf{Desire}}\\ \hline
\multicolumn{4}{c}{\textbf{1st Round Conversation}}\\
\hline
\textbf{P1}: Hi there.. we're looking forward to this camping trip. hope you are, too!  Is there anything in particular you guys need/want out of the extra stockpile we've got? & \textit{Build-Rapport, Discover-Preference} & (Not Given) \textit{No information about participant 2’ preference item} & (Not Given) \textit{No information about participant 1’ preference item}\\ \hline

\textbf{P2}: Hello! Which item do you need the most?& \textit{Discover-Preference} &(Not Given) \textit{No information about participant 1’ preference item}& (Not Given) \textit{No information about participant 2’ preference item}\\ \hline

\multicolumn{4}{c}{\textbf{2nd Round Conversation}}\\\hline
\textbf{P1}: Well, we actually really need all three items.. but I know you might, too. So, I want to make a deal that's fair for you.. is there anything in particular your family needs for the trip? & \textit{Describe-Need, Callout-Fairness, Discover-Preference}& (Not Given) \textit{No information about participant 2’ preference item} &(Not Given) \textit{No information about participant 1’ preference item}\\ \hline

\textbf{P2}: Well, there has to be an item that you need the most. If you tell me what you need the most, we can try to make a deal around it. & \textit{Discover-Preference} & (Not Given) \textit{No information about participant 1’ preference item}& (Not Given) \textit{No information about participant 2’ preference item}\\ \hline

\multicolumn{4}{c}{\textbf{3th Round Conversation}}\\\hline
\textbf{P1}: Since the forest is nearby enough, I think we'd be more interested in ensuring there's enough food instead of firewood for my people, I think. & \textit{Describe-Need,\newline No-Need} & (Food, Firewood, Water) \textit{Participant 2’ high preference is food; medium is firewood and low one is water.}& (Food, Water, Firewood) \textit{Participant 1’ high preference is food; medium is water and low one is firewood.}\\ \hline

\textbf{P2}: Hmm. I really need food too. I don't care much for water either. How about I take all 3 firewood, 2 food, and 1 water? & \textit{Describe-Need,\newline No-Need} & (Food, Water, Firewood) \textit{Participant 1’ high preference is food; medium is water and low one is firewood.}& (Food, Firewood, Water) \textit{Participant 2’ high preference is food; medium is firewood and low one is water.}\\ \hline

\multicolumn{4}{c}{\textbf{4th Round Conversation}}\\\hline
\textbf{P1}: Yeah, that's not going to work for me.. I can give you two food if you really need it, but in exchange I'm going to need two of the water and all the firewood. & \textit{Promote-Coordination} & (Firewood, Food, Water) \textit{Participant 2’ high preference is firewood; medium is food and low one is water.}& (Food, Water, Firewood) \textit{Participant 1’ high preference is food; medium is water and low one is firewood.}\\ \hline

\textbf{P2}: I will just take 3 firewood, 1 food, and 1 water. That work for you? You’re close to the forest, you won't need firewood. & \textit{No-Intention} & (Food, Water, Firewood) \textit{Participant 1’ high preference is food; medium is water and low one is firewood.}  & (Firewood, Food, Water) \textit{Participant 2’ high preference is firewood; medium is food and low one is water.} \\ \hline
\multicolumn{4}{c}{\textbf{...}}\\\hline
\end{tabular}}
\vspace{-0.3cm}
\caption{A negotiation dialogue example. \textbf{P1} and \textbf{P2} represent two participants in this study. The upper part of the table contains three mental state questions in the NegotiationToM benchmark, while the bottom contains annotated label examples. \textcolor{headcolor}{<Agent 1/Agent2>} indicates alternating to query the LLMs for the question regarding agent 1 or agent 2 . \textcolor{tailcolor}{<high/medium/low>} means three individual questions for the agent's high/medium/low preferences on each item. LLMs are required to answer the intention question behind a specific utterance represented by\textcolor{tailcolor}{<Utterance>}.
}
\label{tab:sample-dialogue}
\vspace{-0.6cm}
\end{table*}

\section{NegotiationToM}
Theory of Mind (ToM) describes the ability as humans have to ascribe and infer the mental states of others, and to predict which likely actions they are going to take~\cite{apperly2010mindreaders}. Therefore, it is critical to acquire negotiation strategies based on one's own desire and temporary belief built upon the information presented in conversation. Nevertheless, understanding the Theory of Mind (ToM) inherent in a negotiation dialogue is challenging due to its intricate linguistic features and complex reasoning attributes. Therefore, some considerations have to be taken into account when constructing the NegotiationToM. 

\subsection{Design Considerations for NegotiationToM}
There are several essential design considerations we go through when constructing the NegotiationToM. (1) the scenario of the dataset should be grounded in a human-to-human real-world negotiation (e.g., real-world camping scenario). (2) the dataset should be a natural conversational dataset instead of generated from a synthetic template to avoid reporting bias~\cite{gordon2013reporting} and shortcuts~\cite{DBLP:journals/air/AruLCV23}. (3) the dataset should be equipped with abundant and diverse linguistic features and semantic context (e.g., negotiation argument) instead of bargaining on the numerical value or meaningless counter-offer~\cite{lewis2017deal,he2018decoupling}. (4) the dataset should be ensured to mitigate the risk of potential contamination.

\subsection{CaSiNo}
Inspired by several considerations above, CaSiNo~\cite{DBLP:conf/naacl/ChawlaRCLMG21} was employed as the source data and modified to construct NegotiationToM. CaSiNo is a bilateral human-to-human natural conversational dataset that covers rich linguistic features and many realistic aspects of negotiations, such as small talk, preference elicitation, emotional expression, and convincing strategies based on individual desire. In this dataset, the participants take the role of campsite neighbors and negotiate for food, water, and firewood packages for their upcoming trip. For each conversation, participants discuss individual needs by making various convincing arguments from their camping experiences, such as \textit{Personal Care, Recreational, Group Needs, or Emergency Requirements}. One example of \textit{Group Needs} is "I need more firewood due to having several people join on the trip and needing a bigger fire overall." We illustrate some of these arguments in Table~\ref{tab:sample-reasons} in Appendix~\ref{NegotiationToM_Intentions_details}. Therefore, crafting our benchmark from the CaSiNo offers a range of scenarios based on how to align the preferences of the two parties to reveal more interesting behavior. Furthermore, we present the verification method employed to alleviate the risk of potential contamination within the CaSiNo dataset and demonstrate that this dataset is unlikely to encounter the contamination issue, as detailed in Appendix~\ref{Verification_Contamination}.

\subsection{Theory of Mind in NegotiationToM}
In NegotiationToM, as shown in Table~\ref{tab:sample-dialogue}, it is fundamentally a desire-matching scenario surrounding the item preference order that requires two participants to directly or indirectly align their preference order of item (desire) and adopt corresponding strategies to strive for more high-preference items based on the holding belief (i.e., the assumption of their opponent's item preference order according to the information received in the conversation). Therefore, inspired by the Belief-Desire-Intention (BDI) agent modeling method~\cite{bratman1987intention}, three mental states (i.e., desire, belief, and intention) were employed to evaluate the LLMs' performance in NegotiationToM. All questions about these three mental states are displayed in Table~\ref{tab:sample-dialogue}.

\paragraph{Desire.}
Desires are motivational states that do not necessarily imply commitment, though they usually affect actions~\cite{malle2001distinction,kavanagh2005imaginary}. Unlike beliefs, desires are neither right nor wrong; they are fulfilled or unfulfilled~\cite{searle1983intentionality}. In NegotiationToM scenarios, the desire of the participants is the need for their item preference order, whether they are satisfied or not during the negotiation, and their desire order is the preference order of items. 
Hence, we create a desire question to assess whether the large language model comprehends the desire order of negotiation participants behind each round dialogue with previous conversation history. There are two types of desire order, and one is the global desire order inherently assigned to each participant before the beginning of the negotiation in CaSiNo. Another one is local desire order, which focuses on the local item preference order information behind each round of dialogues and previous conversation history, illustrated in Table~\ref{tab:sample-dialogue}. This local desire order is utilized to form desire questions in NegotiationToM.

\paragraph{Belief.}
Belief refers to a mental state in which an individual assumes a specific stance, attitude, or opinion toward a proposition. In contemporary discussions within the field of philosophy of mind, the term “belief” is commonly employed to denote an individual’s cognitive stance or acceptance of something as true or holding it to be the case~\cite{turiel2008development}. Note that this notion of belief does not inherently require active reflection, nor does it necessitate truthfulness~\cite{armstrong1973belief, moses1993young}. In NegotiationToM, understanding the state of the opponent's item preference order, which is explicitly or implicitly expressed in the conversation, is the main way to form the belief. Therefore, the belief question will query the LLMs on what one participant thinks of another participant's item preferences, given the current round of dialogue with previous conversation history. 

\paragraph{Intention.}
Intention is a mental state formed through rational planning (i.e., negotiation strategy in a negotiation scenario) toward a goal based on the desires and beliefs of the agent. Intentions have been extensively explored in psychology tests, e.g., action prediction~\cite{malle2001distinction} and intention attribution to abstract figures~\cite{castelli2006valley}. Normally, a negotiation strategy is highly associated with corresponding concrete intentions~\cite{belmondo2015negotiating}. Thus, in NegotiationToM, we collect the annotated negotiation strategies from the CaSiNo dataset and map the intentions according to the definition of various strategies. The mapping table is shown in Table~\ref{tab:map_intention}, and the strategy definition and examples are illustrated in Table~\ref{tab:strategies_defin} and~\ref{tab:ann-stats} in Appendix~\ref{NegotiationToM_Intentions_details}. Within our framework, as both \textit{Self-Need} and \textit{Other-Need} are associated with \textit{"Intents to describe a need for an item"} intention, we combine these two strategies into one intention class. 

\begin{table}[t]
\centering
\scalebox{0.6}{
\begin{tabular}{p{4cm}|p{7cm}}
\hline
\multicolumn{1}{c}{\textbf{Strategies}} & \multicolumn{1}{c}{\textbf{Intentions}} \\ 
\hline

Small-Talk   & Intents to build a rapport with the opponent \textit{(Build-Rapport)}\\\hline
Empathy      &  Intents to show empathy \textit{(Show-Empathy)}\\\hline
Coordination & Intents to promote coordination \textit{(Promote-Coordination)}\\ \hline
Elicit-Pref  & Intents to discover the preference order of the opponent \textit{(Discover-Preference)} \\ \hline
Undervalue-Partner    &  Intents to undermine the requirements of their opponent \textit{(Undermine-Requirements)} \\\hline
Vouch-Fairness        &  Intents to callout to fairness \textit{(Callout-Fairness)}\\ \hline
Self-Need/Other-Need  &  Intents to describe a need for an item \textit{(Describe-Need)}\\\hline
No-Need               &  Intents to point out they do not need an item \textit{(No-Need)}\\ \hline
Non-strategic         &  No clear intention in the utterance \textit{(No-Intention)}  \\ \hline
\end{tabular}
}
\vspace{-0.3cm}
\caption{\label{tab:map_intention}Utterance-level intention mapping from the negotiation strategies. The abbreviations of each intention are in brackets.The definition of negotiation strategies and example are in Table~\ref{tab:strategies_defin} and~\ref{tab:ann-stats} in Appendix~\ref{NegotiationToM_Intentions_details}.}
\vspace{-0.5cm}
\end{table}

\subsection{Annotation \& Statistics}
\paragraph{Source data.}
NegotiationToM is annotated based on a multi-turn negotiation dialogue corpus, the CaSiNo \cite{DBLP:conf/naacl/ChawlaRCLMG21} dataset.
Each instance in CaSiNo is an $N$-round alternating dialogue $D_N=[u^a_1, u^b_1, u^a_2, u^b_2, \cdots, u^a_N, u^b_N]$ between two participants, $a$ and $b$~\footnote{When the dialogue ends with user $a$, $u_N^b$ is an empty utterance.}. 
They take on the roles of campsite neighbors and negotiate for \textit{food}, \textit{water}, and \textit{firewood} packages for their upcoming trip.
We adopt the subset with strategy annotations and undertake the annotation on the desire and belief states behind each utterance.

\paragraph{Curating NegotiationToM.} 
The intention state of both participants has already been introduced and mapped from the strategy annotations in CaSiNo. 
We conduct an expert annotation to annotate the beliefs and desires of the two participants in each dialogue (i.e., the perceived preference ranking among \textit{food, water}, and \textit{firewood}).
We recruited five workers who were graduate students in English-speaking universities to conduct the annotation. 
For each dialogue $D_N$, let $D_k$ be the truncated dialogue until round $k$: $D_k=[u_1^a, u_1^b, \cdots, u_k^a, u_k^b]$. 
Then, we ask the workers to annotate the perceived preference ranking for both participants $a$ and $b$ for truncated dialogue $D_k$ ($k\in\{1, 2, \cdots, N\}$) given all $k$ rounds of historical dialogue.
To ensure the annotation quality, we evaluate the workers during the first 100 rounds of conversations and explain their typical errors to them in detail.
More details of the annotation process are in Appendix~\ref{data_annotation_template}.
Although annotating NegotiationToM requires understanding complex dialogues in CaSiNo, we observed high inter-annotator agreement. The Fleiss’s $\kappa$ is 79.03\% \cite{fleiss1971measuring} for NegotiationToM benchmark, the breakdown computation of $\kappa$ are shown in Table~\ref{tab:kappa}.

\begin{table}[t]
\small
\centering
\setlength\tabcolsep{4pt}
\scalebox{0.85}{
\begin{tabular}{l|c}
\toprule
\multicolumn{1}{c|}{\textbf{Task}} & \textbf{Fleiss’s Kappa(\%)}\\
\midrule
Desire (High)     & 83.02  \\
Desire (Medium)   & 72.23  \\
Desire (Low)      & 79.32   \\
\midrule
Belief (High)     & 85.25   \\
Belief (Medium)   & 74.03  \\
Belief (Low)      & 78.81   \\
\midrule
\end{tabular}
}
\vspace{-0.45cm}
\caption{
Inter-rater agreement in terms of Fleiss’s $\kappa$ on belief and desire states.
}
\vspace{-0.5cm}
\label{tab:kappa}
\end{table}

\paragraph{Statistics.}
NegotiationToM contains 395 dialogues with 2,380 rounds of conversations (truncated dialogues) and 4,618 utterances. 
Each utterance has seven questions and annotated labels, including three designed sub-questions for both belief and desire states (i.e., high/medium/low preference items) and one tailored question for intention. There are a total of 13.8 thousand questions, and the detailed statistics and comparison with contemporary ToM datasets are shown in Table~\ref{tab:dataset_stats} in Appendix~\ref{data_annotation_template}.

\begin{table*}[!t]
\small
\centering
\setlength\tabcolsep{4pt}
\scalebox{0.83}{
\begin{tabular}{l|c|c|c|c|c|c|c}
\toprule
\multicolumn{1}{c|}{\multirow{2}{*}{\textbf{Model}}} & \multicolumn{1}{c|}{\textbf{Desire}} & \multicolumn{1}{c|}{\textbf{Belief}} & \multicolumn{2}{c|}{\textbf{Intention}} & \multicolumn{1}{c|}{\textbf{\textit{All}}} & \multicolumn{2}{c}{\textbf{Consistency}}\\

                          & Exact.Match.(\%)    & Exact.Match.(\%)      & Micro.F1(\%)     & Macro.F1(\%)  & Exact.Match(\%) &  Desire(\%)  & Belief(\%) \\
\midrule
LLaMa2-Chat(13B)          & 15.41  & 14.63  & 22.66    & 19.82    & 0.56      & 0.76   & 0.76        \\
LLaMa2-Chat(13B) (CoT)     & 16.15   & 18.21  & 24.20   & 20.81    & 0.61     & 0.76   & 0.76          \\
LLaMa2-Chat(13B) (Few-Shot)    & 13.49   & 12.54  & 26.30 & 21.76  & 0.64 & 0.90   & 0.80 \\
LLaMa2-Chat(70B)          & 24.40   & 21.58  & 33.23   & 27.70    & 0.45      & 1.78   & 1.51      \\
LLaMa2-Chat(70B) (CoT)     & 30.34   & 24.23  & 30.57   & 26.26    & 1.06    & 2.28   & 0.00          \\
LLaMa2-Chat(70B) (Few-Shot)    & 26.95   & 22.84  & 35.77 & 28.10  & 1.28 & 1.32   & 0.91 \\
\midrule
Claude-v1.3               & 26.27   & 23.15  & 30.80   & 27.81    & 1.50     & 0.25   & 1.01          \\
Claude-v1.3 (CoT)         & 44.63   & 37.18  & 31.12   & 28.25    & 1.62     & 4.81   & 1.52        \\
Claude-v1.3 (Few-Shot)    & 30.73   & 30.68  & 32.35 & 30.10    & 1.80 & 3.23   & 1.20 \\
Claude-v2.1               & 45.10   & 39.49  & 37.48   & 32.94    & \underline{3.40} & 6.08   & 3.54 \\
Claude-v2.1 (CoT)         & 50.13   & 40.52  & \textbf{39.93}   & \textbf{35.67}    & \textbf{3.68}   & 6.07   & 4.05\\
Claude-v2.1 (Few-Shot)    & 48.77   & 41.88    & \underline{38.23} & \underline{34.32}   & 2.90 & 6.25   & 4.28 \\
\midrule
ChatGPT 0613              & 18.60   & 13.04  & 33.95   & 29.73   & 0.43    & 0.00  & 0.00        \\
ChatGPT 0613 (CoT)         & 28.45   & 21.00  & 36.71    & 30.79    & 0.78  & 0.76 & 0.25       \\
ChatGPT 0613 (Few-Shot)    & 19.24  & 17.02    & 36.29 & 30.84    & 2.16  & 0.00  & 0.00 \\

GPT-4 0613                & 62.77   & \underline{57.62} &  29.84  & 27.15    & 2.58  & 13.67 & 10.63  \\
GPT-4 0613 (CoT)           & \textbf{63.29}   & \textbf{58.18} & 34.90   & 31.26    & 2.79       & \textbf{17.72}   & \textbf{14.18}      \\
GPT-4 0613 (Few-Shot)      & \underline{62.89}   & 52.08 & 35.10  & 33.21    & 2.51 & \underline{15.94}   & \underline{12.76}\\
\midrule
Human                & 91.14   & 91.14  & 83.75 & 84.65   & 43.78  &  75.44 &  75.44\\

\midrule
\end{tabular}
}
\vspace{-0.4cm}
\caption{
Main results of models for the NegotiationToM. The best results are \textbf{bold-faced}, and the second-best ones are \underline{underlined}. The conversation consistency (\%) of the models’ responses for answering correctly in whole dialogues. All models received zero consistency scores in the intention aspect, as the intention mental state owned a multi-label in an utterance and imposed difficulties to generate exact match labels in the whole label.
}
\label{tab:experiment}
\end{table*}

\section{Experimental Setting}

\subsection{Baseline Models}
In this work, we test six recent instruction-tuned large language models: GPT-4 \cite{DBLP:journals/corr/abs-2303-08774}, ChatGPT \citep{openai2022chatgpt}, Claude-v1.3\footnote{https://www.anthropic.com/news/introducing-claude}~\cite{IntroducingClaude1}, Claude-v2.1\footnote{https://www.anthropic.com/news/claude-2-1}~\cite{IntroducingClaude2}, Llama-2 Chat 13B~\cite{DBLP:journals/corr/abs-2307-09288}, and Llama-2 Chat 70B~\cite{DBLP:journals/corr/abs-2307-09288}. Descriptions for each model are in Appendix ~\ref{sec:Baseline_models}. By following the common practices in the theory of mind field~\cite{DBLP:conf/emnlp/0002SZ0K0S23,DBLP:journals/corr/abs-2306-15448, DBLP:conf/acl/ShapiraZG23}, we test these models with two types of prompts: (1) one is zero-shot prompting and we utilize the prompt template in \citet{DBLP:conf/iclr/RobinsonW23} to formulate the task as a multiple choice question answering problem as a baseline. (2) another one is the Chain-of-Thought (CoT) prompting method by following~\cite{DBLP:conf/nips/KojimaGRMI22} and using the prompt “let’s think step by step.” Apart from these two settings, we also assess the LLMs' performance by using the few-shot setting to validate whether LLMs can improve their performance with input-output exemplars. We concatenate four tailored exemplars for desire and belief states and seven designed exemplars for intentions, covering all the input-output exemplars' labels. More configuration details can be found in Appendix~\ref{sec:hyparameter}, and the prompt template refers to Appendix~\ref{sec:Prompt_Template}. To measure the specific performance gap between humans and the state-of-the-art machine on the NegotiationToM, we employ three graduate students in computer science to complete the human evaluation task. More details of human evaluation are shown in Appendix~\ref{sec:Appendix_for_huamn_performance}.

\subsection{Metrics}
We report the exact match percentages of all three high, medium, and low preferences for desire and belief classification, and only both of these three preferences that answer correctly count toward correct. The micro F1 score and macro F1 score are reported for the multi-label intention classification by following prior works~\cite{DBLP:conf/aaai/HouLWCL21, DBLP:conf/emnlp/WuSJ21, DBLP:conf/acl/MogheRGVKB23, DBLP:conf/emnlp/VulicCSMWB22}. Moreover, we report the \textit{“All”} score, which requires the models to answer correctly all three ToM question types, which include desire, belief, and intention for the same information piece in the conversation. Furthermore, we report the consistency score, which requires the models to answer ToM questions correctly for the whole negotiation conversation. This metric aims to measure how well the models show consistent understanding and track the agent's mental state change throughout the whole conversation.


\section{Experimental Result}
\subsection{Main Result}
Table~\ref{tab:experiment} summarizes the main results of the state-of-the-art large language models in NegotiationToM, from which we derive the following conclusions. \textbf{First}, the performance of all models is significantly worse than human performance, even after employing the zero-shot chain-of-thought (CoT) method. There is a significant performance gap between machines and humans in the ToM negotiation evaluation scenario. Specifically, compared with the best state-of-the-art models in each mental state, the performance gap is 27.85\% in desire, 32.96\% in belief, 43.82\% in Micro F1 score, and 48.98\% in Macro F1 score in intention. \textbf{Second}, GPT-4 0613 (CoT) achieves the best performance among all the models regarding inferring desire and belief, while Claude-v2.1 (CoT) outperforms all other models in the intention classification task in NegotiationToM. \textbf{Third}, we observe that most models received an improvement in scores when the chain-of-thought (CoT) method was applied. Nevertheless, there are still significant score gaps compared to human performance.

\paragraph{Few-Shot Performance} Although the few-shot setting is not a common practice, we also include it to see whether it performs better than the other two settings. The results illustrated in Table~\ref{tab:experiment} display that the few-shot setting of most models gains an improvement over the zero-shot setting but is worse than the CoT prompting method in NegotiationToM. Interestingly, some models (e.g., GPT-4) with few-shot exemplars received a better performance than the CoT method in intention states.

\subsection{Large Language Model on \textit{All} score}
To fully assess the ToM capability of large language models to understand other's mental states in each round of dialogues, we report the \textit{"All"} score in Table~\ref{tab:experiment}. This metric required the machine equipped with various ToM abilities to correctly answer three mental states (i.e., desire, belief, and intention) under the same information piece in the conversation. The Claude-v2.1 (CoT) outperforms all other large language models and receives a 3.68\% in this \textit{all} metric. It may be attributed to the exceptional intention ToM ability of Claude-v2.1 (CoT), but it also obtains a relatively high performance on the desire and belief aspects of ToM. However, it is worth mentioning that the performance of the machine on the \textit{all} metric is far away from human performance, which is 43.78\%.

\subsection{How Well Large Language Model on Tracking Mental States Change in Conversation}
To assess how well large language models can track the mental states of negotiation participants in conversations and coherent understanding of others’ mental states with increased available information. Thus, it is crucial to evaluate the consistency and faithfulness of the large language model for the conversation context of the whole theory of mind-based dialogue. The consistency score is presented in Table~\ref{tab:experiment}, GPT-4(CoT) received an excellent performance on this metric compared with other models (e.g., Claude-v2.1 (CoT)), which are 17.72\% in desire and 14.18\% in belief. Nevertheless, there is a huge performance gap between machines and humans in this consistency metric, demonstrating LLMs still lack of ability to track the mental state change during the conversation. It is noted that all models received zero consistency scores in the intention aspect, as the intention mental state owned a multi-label in an utterance and imposed difficulties to generate exact match labels in the whole conversation.



\begin{table}[!t]
\small
\centering
\setlength\tabcolsep{4pt}
\scalebox{0.9}{
\begin{tabular}{l|c|c}
\toprule
\multicolumn{1}{c|}{\multirow{2}{*}{\textbf{Model}}} & \multicolumn{2}{c}{\textbf{Question Forms}}\\
                          &  Desire   & Belief \\
\midrule
ChatGPT 0613 (ranking form)      & 2.88  & 9.24  \\ 
ChatGPT 0613 (individual form)  & 9.18 & 11.7 \\ 
ChatGPT 0613 (combined form)     & \textbf{18.60} & \textbf{13.04}   \\
\midrule
GPT-4 0613 (ranking form)      & 20.10  & 16.80  \\ 
GPT-4 0613 (individual form)  & 40.01 & 36.88 \\
GPT-4 0613 (combined form)     & \textbf{62.77} & \textbf{57.62}   \\
\midrule
\end{tabular}
}
\vspace{-0.4cm}
\caption{
The zero-shot performance of three question types. The intention task is ignored in this experiment as this task in NegotiationToM is a multi-label classification.  
}
\vspace{-0.5cm}
\label{tab:question_forms}
\end{table}

\subsection{The Effect of Question Format} \label{Question_Forms}
With the performance of LLMs varying significantly due to the sensitivity of prompt templates~\cite{DBLP:conf/naacl/WebsonP22}, we assessed the performance of two state-of-the-art models, ChatGPT and GPT-4, to study the effect of various question formats on their performance. In this experiment, we adopt three types of question formats, including \textit{\textbf{ranking format, individual format, and combined format}} for desire and belief mental state. The \textit{\textbf{combined format}} is the baseline prompt template adopted in our main experiment, which combines all three questions regarding the preference order into a single question and asks the LLMs to answer it simultaneously. The \textit{\textbf{ranking format}} indicates collecting high, medium, and low preference items as one ranking answer. The \textit{\textbf{individual format}} splits the high, medium, and low preference questions into three questions and feeds them to LLMs individually. The combined, ranking, individual, question format prompt template is shown in Tables~\ref{tab:baseline_prompt_template}, \ref{tab:ranking_template},~\ref{tab:individual_template_1}, and \ref{tab:individual_template_2} in Appendix~\ref{sec:Prompt_Template}. 

The performance shown in Table~\ref{tab:question_forms} demonstrates that the question format indeed affects the LLMs' performance, and the combined format performs better than other formats. It may result from the combined format imposing the constraint for LLMs to avoid answering some unreasonable and implausible response. After the case study on error cases from the GPT-4 with individual form, we find that it is more challenging for models to combine with different types of reasoning while conducting the theory of mind reasoning. For example, when the GPT-4 may correctly answer that the agent's highest preference item is water and the lowest one is food, the model may randomly answer medium preference as there is no information of medium preference provided in conversation. Models cannot answer the medium preference of firewood (there are only three items) because they cannot effectively adopt deductive reasoning ability when performing theory of mind reasoning. Other models (e.g., ChatGPT) also suffer from this issue more seriously, although the combined format slightly mitigates this issue to some extent.

\begin{table}[!t]
\small
\centering
\setlength\tabcolsep{4pt}
\scalebox{0.65}{
\begin{tabular}{l|c|c}
\toprule
\multicolumn{1}{c|}{\multirow{2}{*}{\textbf{Model}}} & \multicolumn{2}{c}{\textbf{Intention}}\\

                           & Micro.F1(\%)     & Macro.F1(\%) \\
\midrule
LLaMa2-Chat(13B)               & 13.13 & 10.53              \\
LLaMa2-Chat(13B) (w B.D.)      & 18.61 & 15.44              \\
LLaMa2-Chat(70B)               & 22.70 & 18.41          \\
LLaMa2-Chat(70B) (w B.D.)      & 25.94 & 21.18               \\

\midrule
Claude-v1.3                   & 21.77	& 18.52                \\
Claude-v1.3 (w B.D.)          & 26.95   & 23.35      \\
Claude-v2.1                   & 27.12	& 24.48    \\
Claude-v2.1 (w B.D.)          & \textbf{34.56}   & \textbf{30.02}       \\
\midrule
ChatGPT 0613                  & 23.42  & 18.44              \\
ChatGPT 0613 (w B.D.)         & 28.99  & 25.93              \\
GPT-4 0613                    & 26.31  & 23.86       \\
GPT-4 0613 (w B.D.)           & \underline{32.71}  & \underline{29.77}    \\
\midrule
\end{tabular}
}
\vspace{-0.4cm}
\caption{
Results of models for the CaSiNo strategy prediction. \textit{w B.D.} indicate with the input with desire and belief. The best results are \textbf{bold-faced}, and the second-best ones are \underline{underlined}.
}
\label{tab:casino_strategy}
\end{table}

\subsection{CaSiNo Negotiation Strategy Prediction}
To validate the significance of our annotated desire and belief states, we append the information from these two states into the prompt template and assess whether it enhances the model performance on the negotiation strategy prediction task from the CaSiNo dataset. The baseline prompt template and the prompt template with belief and desire are shown in Tables~\ref{tab:strategy_prediction_template} and~\ref{tab:strategy_prediction_template_1} in Appendix~\ref{sec:Prompt_Template}. The micro F1 score and macro F1 were employed as this task metric, and the result is reported in Table~\ref{tab:casino_strategy}. All models incorporating the information from these two mental states received a significant improvement over the baselines. Specifically, by integrating the signals from desire and belief, the Claude-v2.1 model obtained a 7.44\% gain in the micro F1 score and a 5.54\% gain in the macro F1 score. It demonstrates that the effectiveness of our annotated theory of mind states (i.e., desire and belief) helps LLMs to infer the negotiation strategy behind each utterance. For example, with the understanding that agent 1’s preference order of items is \textit{Not Given, Not Given, and Not Given}, and agent 2’s preference order of items is \textit{Firewood, Not Given, and Not Given}. Agent 2 may take the elicit-preference strategy to elicit the preference order of agent 1 for further negotiation.

\begin{figure}[t]
    \centering
    \scalebox{0.5}{
    \includegraphics[width=1\textwidth]{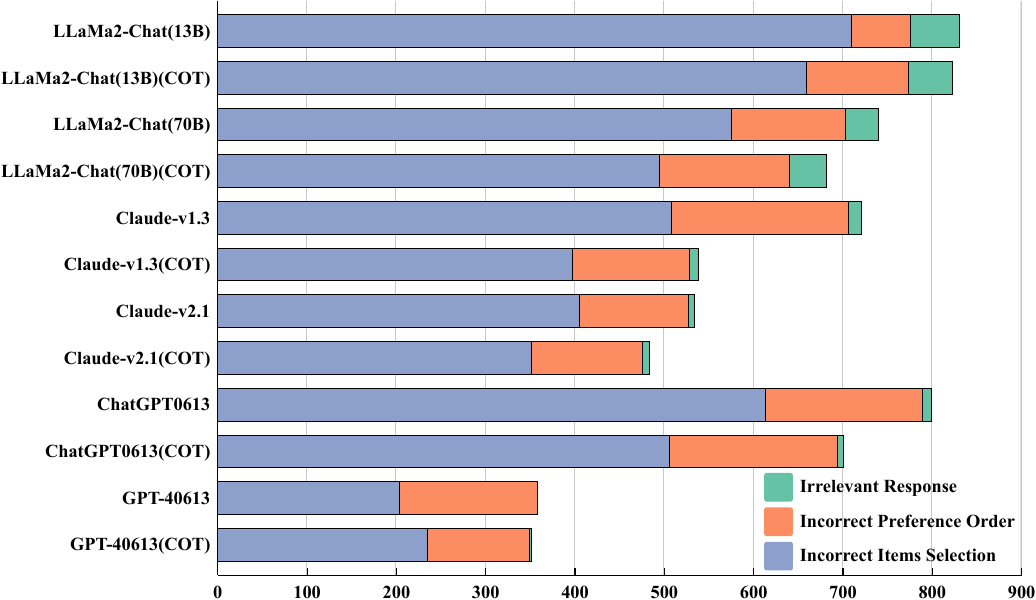}
    }    
    \vspace{-0.7cm}
    \caption{Model errors for desire state}
    \label{fig:error_desire}
    \vspace{-0.4cm}
\end{figure}

\begin{figure}[t]
    \centering
    \scalebox{0.5}{
    \includegraphics[width=1\textwidth]{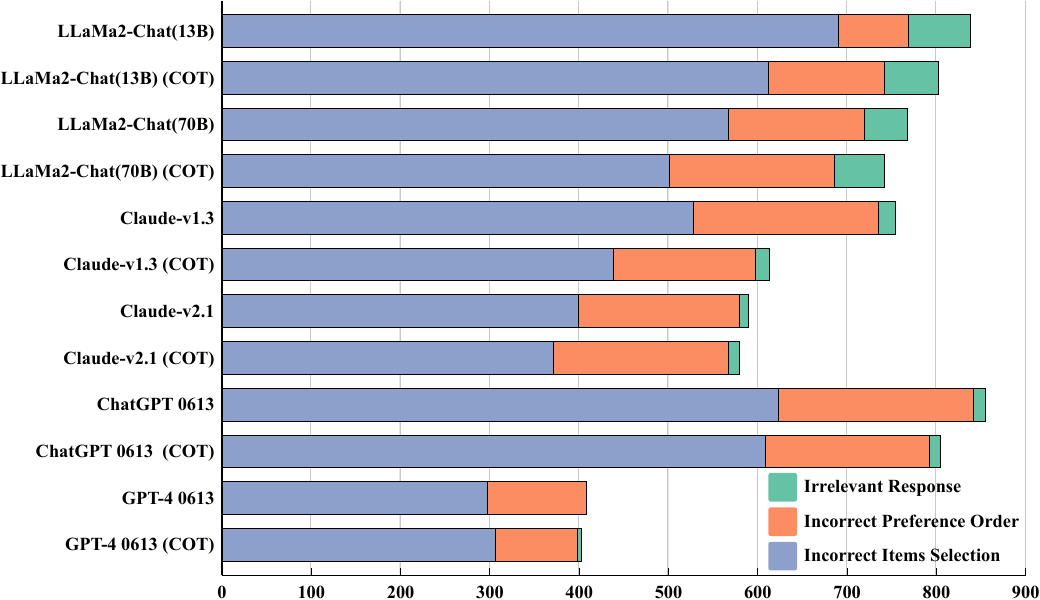}
    }    
    \vspace{-0.7cm}
    \caption{Model errors for belief state}
    \label{fig:error_belief}
    \vspace{-0.5cm}
\end{figure}

\begin{figure}[t]
    \centering
    \scalebox{0.5}{
    \includegraphics[width=1\textwidth]{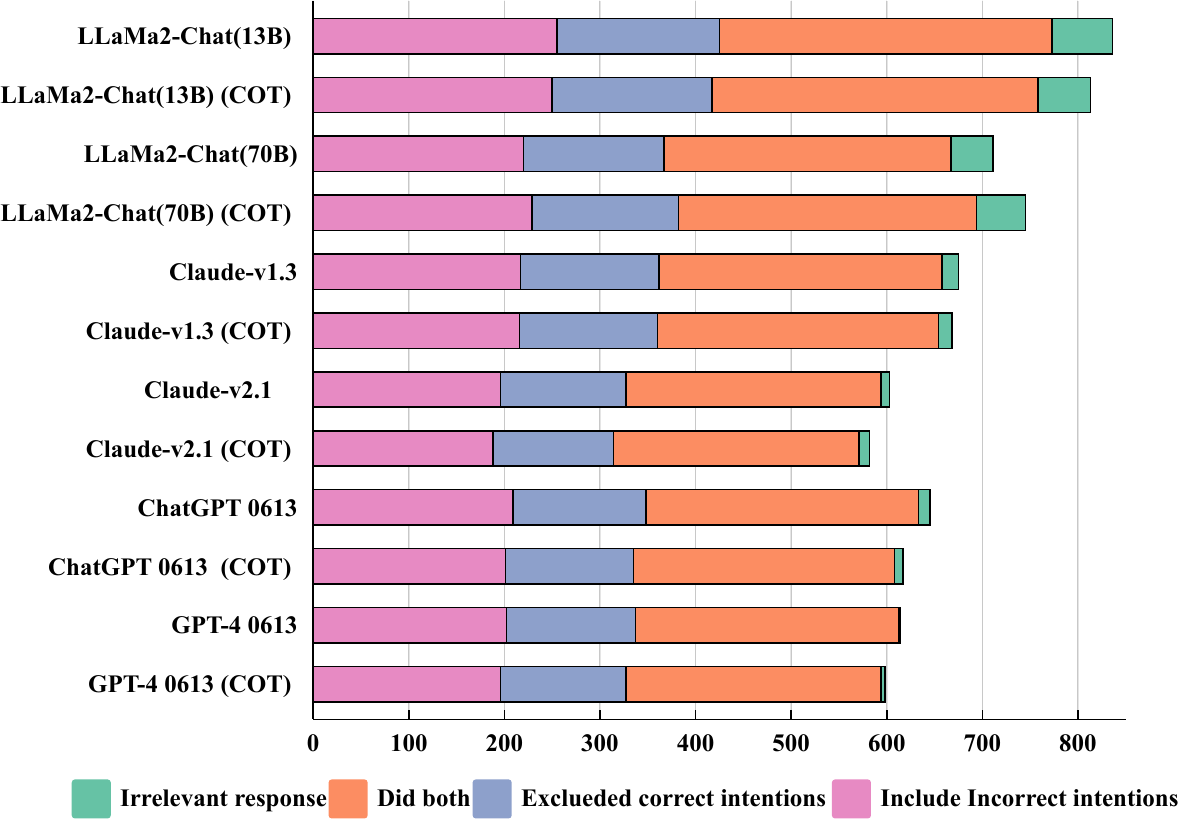}
    }    
    \vspace{-0.6cm}
    \caption{Model errors for intention state}
    \label{fig:error_intention}
    \vspace{-0.3cm}
\end{figure}

\subsection{What Types of Error LLMs Make}
To understand the type of error LLMs make on the NegotiationToM benchmark, we sampled 1,000 LLMs' responses and counted the error categories among them.

\paragraph{Types of Error LLMs make on Desire and Belief State}
Figures~\ref{fig:error_desire} and~\ref{fig:error_belief} summarize the error types of desire and belief state for each model with and without CoT reasoning. 
All models make more errors by including incorrect items and excluding correct items (i.e., Incorrect Items Selection). For example, LLMs tend to select items (e.g., water) randomly rather than answer "Not Given" when there is insufficient information to determine the preferred items. With the CoT method adopted, this error will be decreased for most models as LLM conducted reasoning on the conversation context and tried to explain and respond to a reasonable answer.

\paragraph{Types of Error LLMs make on Intention State}
In terms of intention state, all models without and with CoT tend to select more intention choices, resulting in a high error rate in the "including incorrect intentions" and "did both" (i.e., include incorrect intentions and excluded correct intentions) error types.
Another finding is that LLaMa2 series models respond to many irrelevant responses, such as repeating the questions, and do not raise any relevant answers.

\begin{figure}[t]
    \centering
    \scalebox{0.5}{
    \includegraphics[width=1\textwidth]{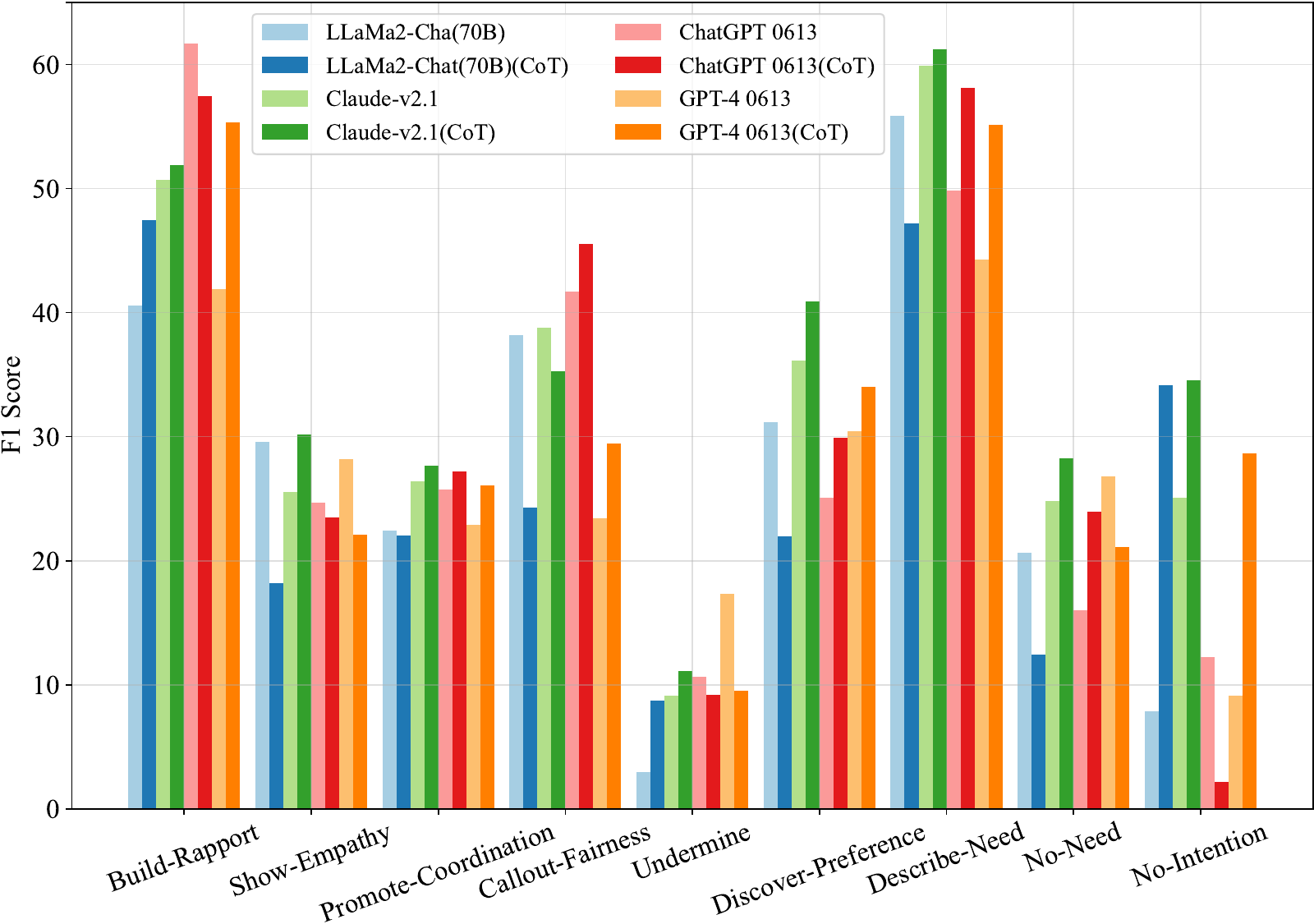}
    }    
    \vspace{-0.8cm}
    \caption{The label-wise intention dimension results of large language models in NegotiationToM. \textit{undermine} stand for the undermine-requirements intention.}
    \label{fig:Intention_Results_labelwise}
    \vspace{-0.5cm}
\end{figure}

\paragraph{Label-wise for Intentions State} To further explore the LLMs' performance on nine intention subclasss in NegotiationToM, Figure~\ref{fig:Intention_Results_labelwise} illustrates the F1 scores (\%) for each subclass. 
The results indicate that LLMs exhibit strong performance in predicting Build-Rapport and Describe-Need intentions, while their performance in predicting "undermine-requirements" and "No-Intention" is poor. Notably, Claude-v2.1 (CoT) outperforms other models in more than half of the subclass in intentions, demonstrating its proficiency in inferring others' intentions. For a detailed subclass result covering all models, please refer to Figure~\ref{fig:Overall_Intention_Results_labelwise} and Table~\ref{tab:label-wise_intention} in Appendix~\ref{sec:Appendix_for_Intention_Result}.

\section{Conclusion}
This work introduces NegotiationToM, a new benchmark designed to stress-test machine ToM in real-world negotiation surrounding covered multi-dimensional mental states. We performed comprehensive and detailed experiments to evaluate LLMs' capability on the NegotiationToM benchmark and discovered that LLMs exhibit inferior performance compared to humans in the NegotiationToM task.

\section*{Limitations}
\paragraph{Passive benchmark to evaluate the ability of LLMs}
Although our benchmark is to stress-test machine ToM ability in negotiation surrounding compassed multi-categories of mental states, NegotiationToM, and existing prior benchmarks are passive benchmarks that primarily adopt a passive observer role to test language agents~\cite{DBLP:conf/emnlp/MaSPC23}. These benchmarks passively assess the ToM ability of LLMs and lack active interaction with and engagement between the agent and other entities involved in the situated environment. The active ToM benchmark should treat the language model as an active agent that perceives the physical and social context, reasons about others’ mental states, communicates with other agents, and interacts with the environment to complete pre-defined tasks. The future work of this paper will employ the language model to act as an agent to actively interact with other model agents by using the Belief-Desire-Intention agent modeling method to generate a rational negotiation strategy. For example, based on the information of desire, belief, and intention, the language model agent will actively acquire a negotiation strategy arguing more benefits or enhancing the cooperation.

\section*{Ethics Statement}
In this work, we conformed to recognized privacy practices and rigorously followed the data usage policy. We declare that all authors of this paper acknowledge the \emph{ACM Code of Ethics} and honor the code of conduct. This paper introduces a benchmark for stress-testing machine theory of mind of large language model on the negotiation surrounding. Our benchmark is modified from the CaSiNo~\cite{DBLP:conf/naacl/ChawlaRCLMG21}, an English-based negotiation dataset. They conducted a data post-processing step for filtering inappropriate language use (e.g., English swear words) dialogues although this situation rarely occurred in the negotiation process. Therefore, we can foresee no immediate social consequences or ethical issues as we do not introduce social/ethical bias into the model or amplify any bias from the data. Moreover, the license CaSiNo CC-BY-4.0 license allows us to modify the data for research, and this fulfills their intended use. 

\section*{Acknowledgements}
The authors of this paper were supported by the NSFC Fund (U20B2053) from the NSFC of China, the RIF (R6020-19 and R6021-20) and the GRF (16211520 and 16205322) from RGC of Hong Kong. We also thank the support from NVIDIA AI Technology Center (NVAITC).

\bibliography{anthology,custom}
\bibliographystyle{acl_natbib}

\newpage
\appendix

\section{Appendix for NegotiationToM} 
\subsection{Verification of Potential Contamination in CaSiNo} \label{Verification_Contamination}
Most of the existing available benchmarks in the NLP field were released prior to the initiation of the LLM training process, indicating that these datasets are likely to have been utilized during the pre-training phase and post-training phase (i.e., SFT~\cite{DBLP:conf/nips/Ouyang0JAWMZASR22} or RLHF~\cite{DBLP:conf/nips/ChristianoLBMLA17}) of LLMs~\cite{DBLP:conf/iclr/GolchinS24, DBLP:conf/aaai/LiF24}. Therefore, we follow the method proposed by~\citet{DBLP:conf/iclr/GolchinS24} to assess the potential contamination issues in the CaSiNo dataset. We sample 100 instances from the CaSiNo dataset and prompting the ChatGPT and GPT-4 to generate the likely next dialogue. After human validation of these 100 instances, none of these generated dialogues corresponded to the original dataset. The prompting template and outputs are illustrated in Table~\ref{tab:verify_casino}. Furthermore, we also test these 100 instances by using the prompt template "Give a CaSiNO negotiation dialogue example and its answer for negotiation strategy." 
The outcome aligns with the prompting template depicted in Table~\ref{tab:verify_casino}, demonstrating that the risk of potential contamination is mitigated; however, no systematic approach can effectively address the contamination issue unless all training datasets utilized for LLMs are made publicly available.

\subsection{NegotiationToM Annotation} \label{data_annotation_template}
\begin{figure*}[h]
    \centering
    \includegraphics[width=0.8\textwidth]{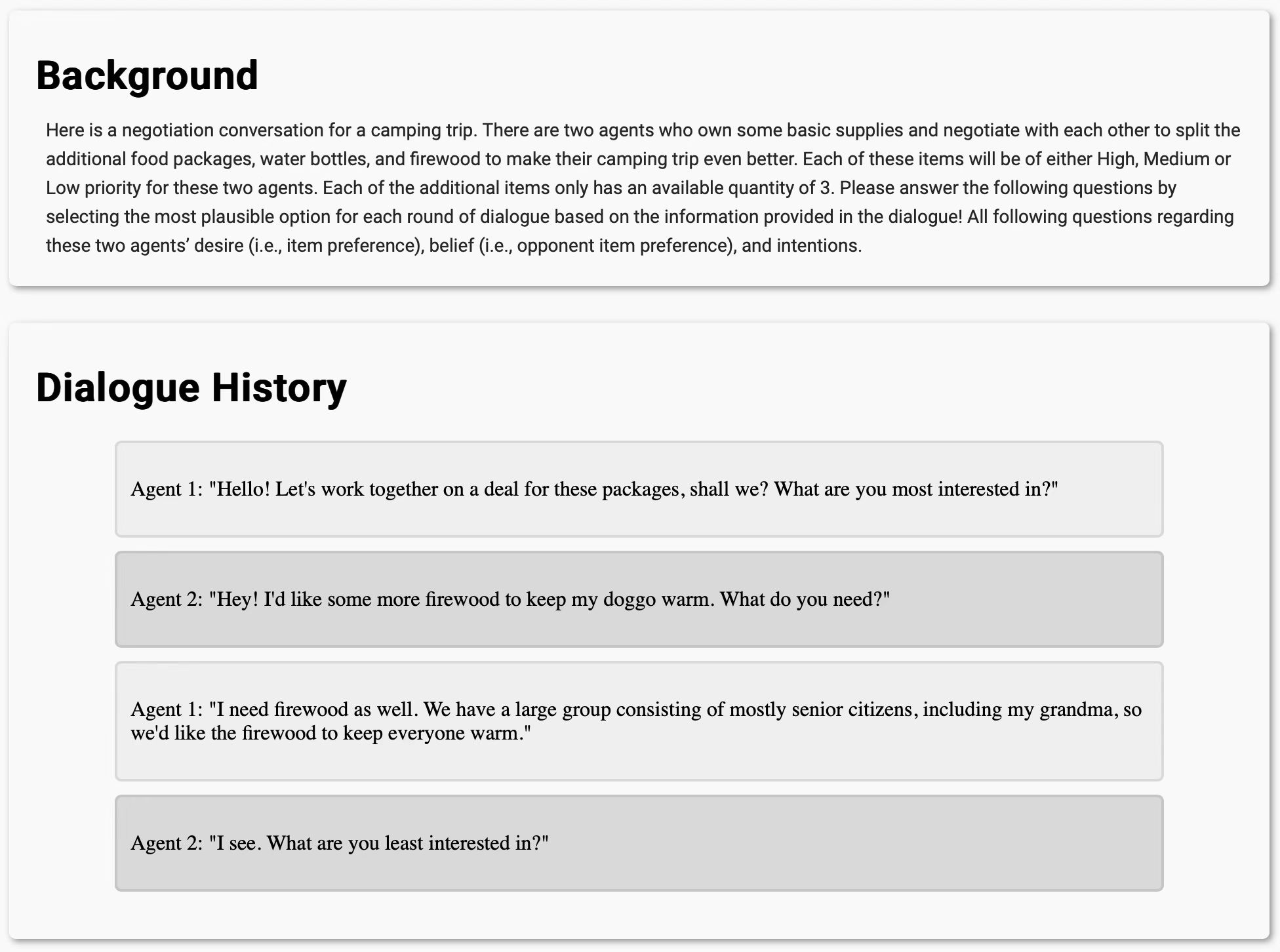}
    \caption{The instructions and background information used for annotation.}
    \label{fig:Annotation_BackGround}
\end{figure*}

\begin{figure*}[h]
    \centering
    \includegraphics[width=0.8\textwidth]{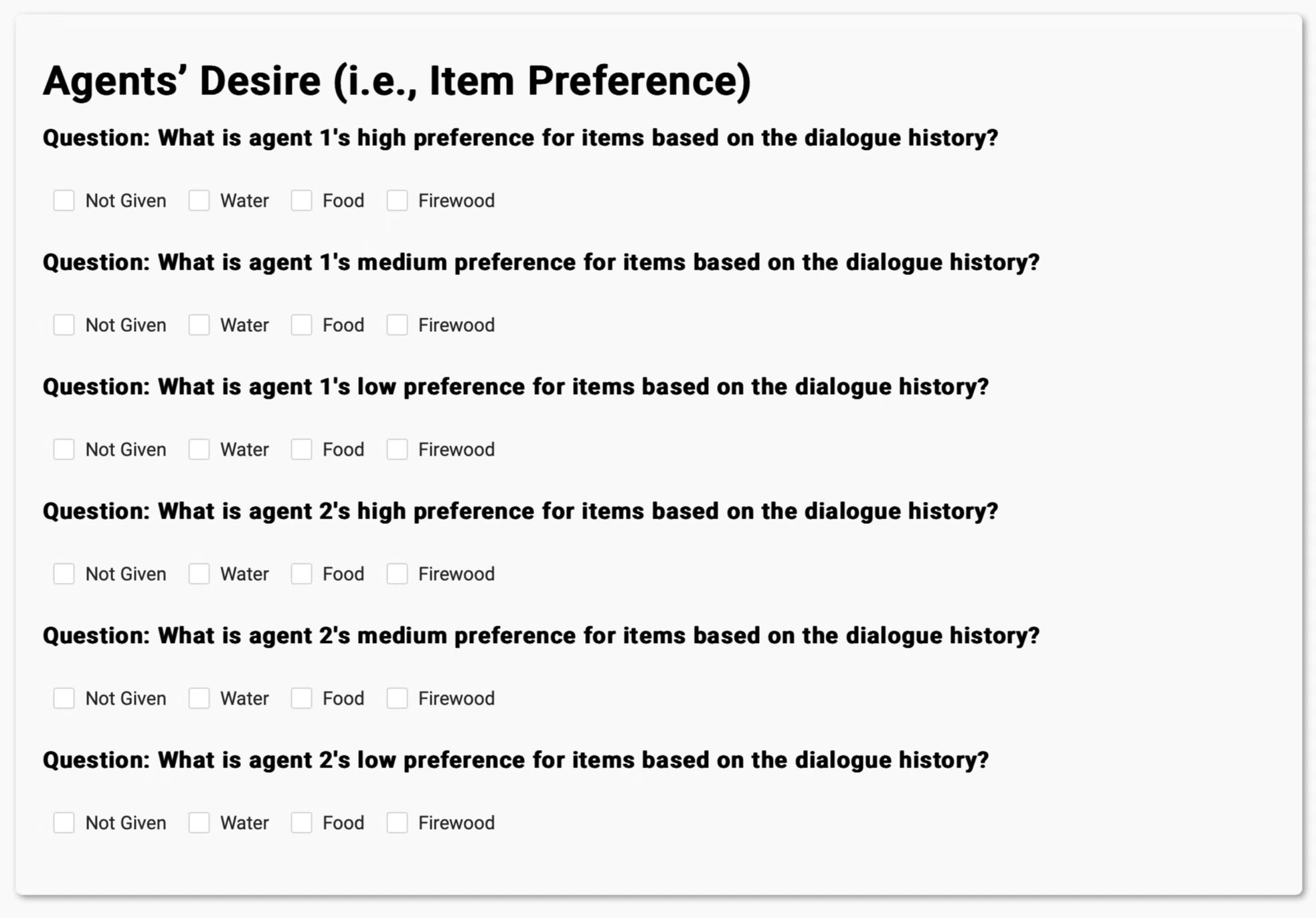}
    \caption{The template for presenting questions regarding the annotation of agent desire.}
    \label{fig:Annotation_Desire}
\end{figure*}

\begin{figure*}[h]
    \centering
    \includegraphics[width=0.8\textwidth]{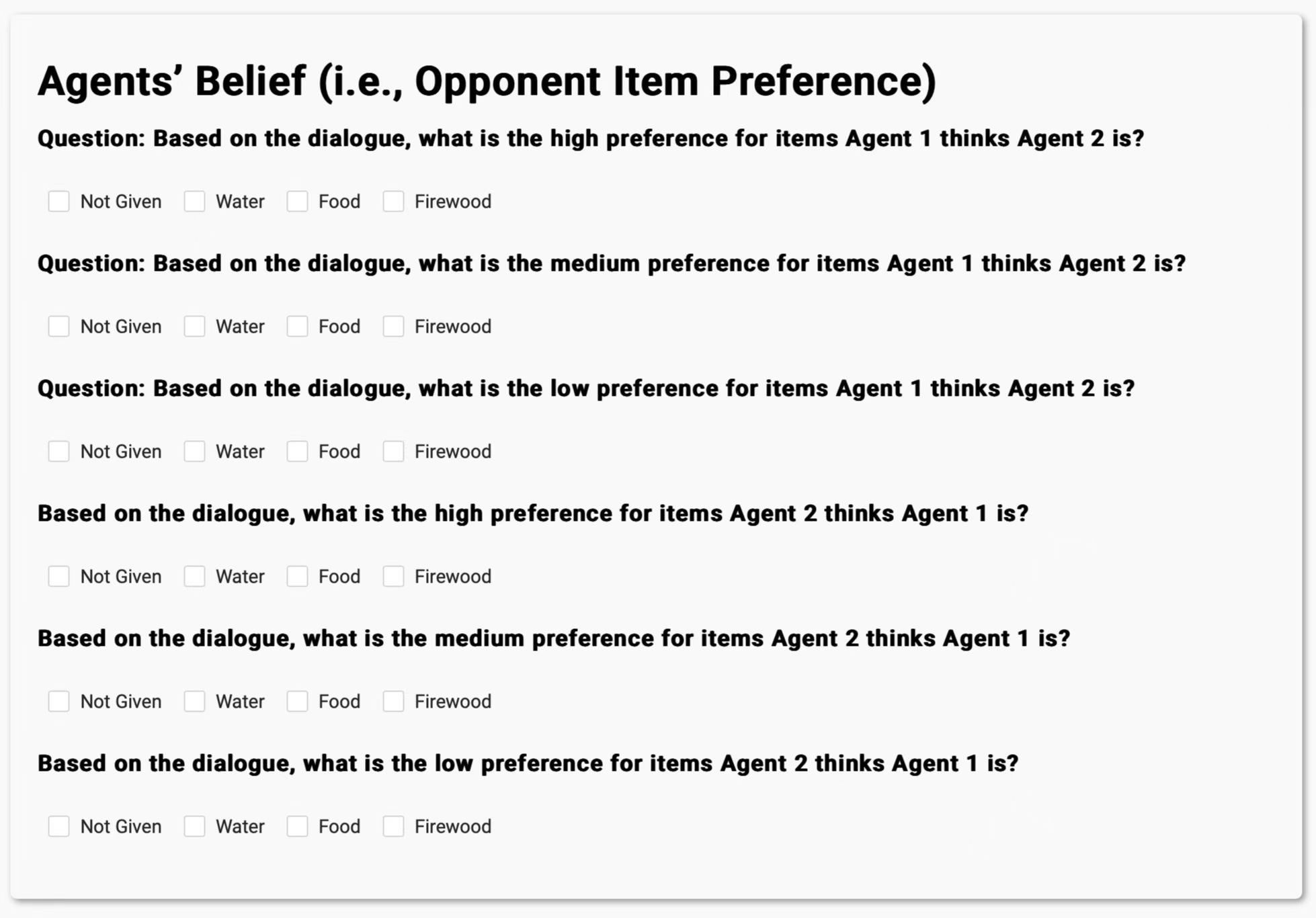}
    \caption{The template for presenting questions regarding the annotation of agent belief.}
    \label{fig:Annotation_Belief}
\end{figure*}

\begin{figure*}[h]
    \centering
    \includegraphics[width=0.8\textwidth]{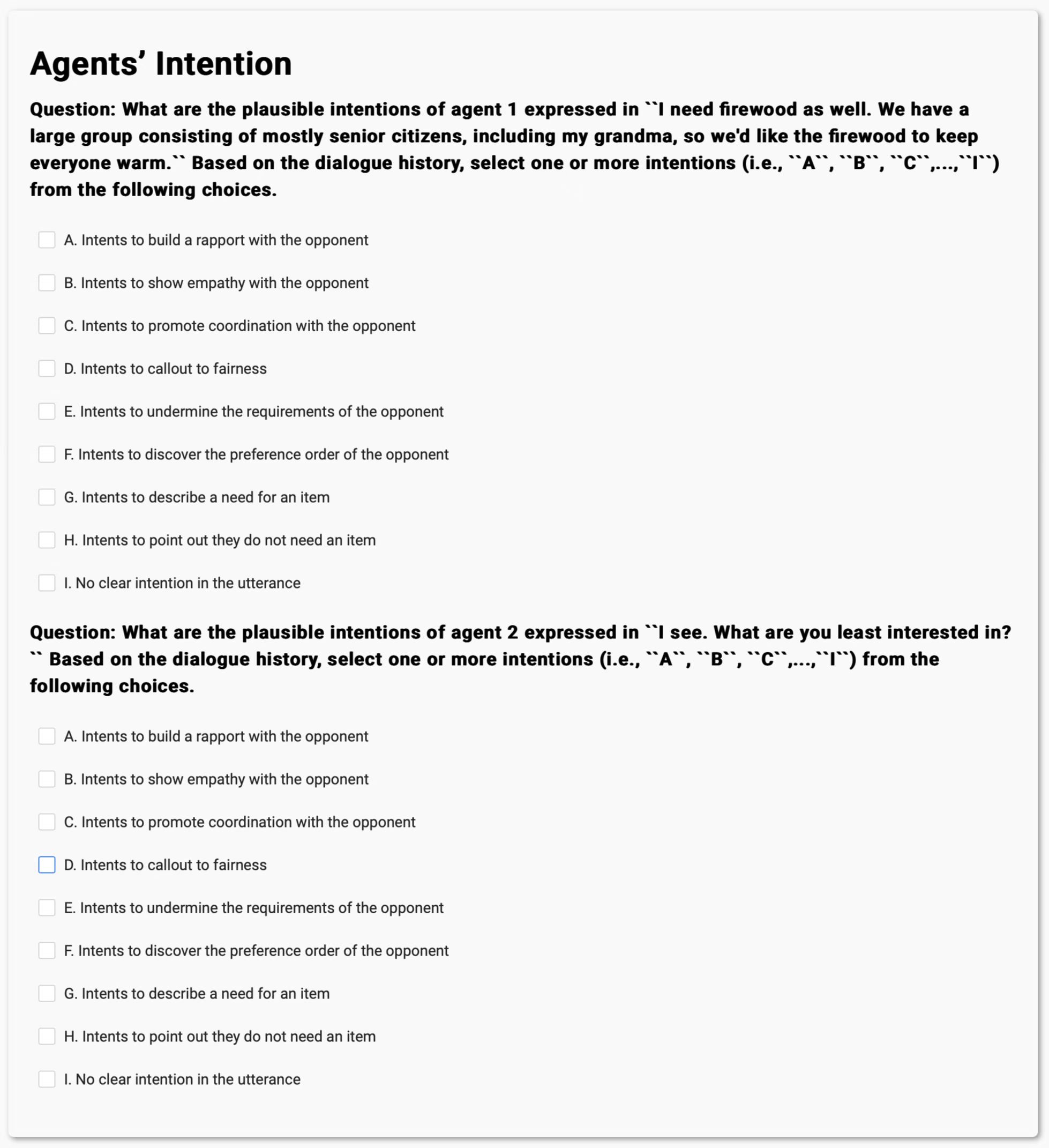}
    \caption{The template for presenting questions regarding the annotation of agent intention.}
    \label{fig:Annotation_Intention}
\end{figure*}

In this section, we showcase our annotation instructions and templates used for annotation. The instructions are used to introduce the background of the negotiation conversation and instruct the workers to perform the annotation based on the dialogue history. The annotation instruction and annotation template are presented in Figure~\ref{fig:Annotation_BackGround}, ~\ref{fig:Annotation_Desire}, ~\ref{fig:Annotation_Belief}, and \ref{fig:Annotation_Intention}. The detailed statistics and comparison with contemporary ToM datasets are shown in Table~\ref{tab:dataset_stats}.

\subsection{Details for NegotiationToM Intentions} \label{NegotiationToM_Intentions_details}
We provide a brief overview of Table~\ref{tab:strategies_defin}, which presents the definitions of various ToM intentions in negotiation strategies. 
These definitions help us understand the intentions of the agents involved in the negotiation process. 
The table offers definitions for strategies such as Small-Talk, Empathy, Coordination, No-Need, Elicit-Pref, Undervalue-Partner, Vouch-Fairness, Self-Need, Other-Need, and Non-strategic. Each definition explains the specific meaning and context of the respective strategy in the negotiation process. By understanding these strategy definitions, we can better comprehend the negotiation interactions between agents and how the intention relates to desire and belief states during the negotiation process.

\paragraph{NegotiationToM Arguments} 
Table~\ref{tab:sample-reasons} shows arguments for various items (i.e., Food, Water, Firewood) in four categories: Personal Care, Recreational, Group Needs, and Emergency. For example, participants may need more food for larger-sized teenage children, and more water for hydration or emergencies. The diversity of negotiation arguments raised by the human participants provided various scenarios for stress-testing LLMs by avoiding shortcuts and spurious correlation issues.

\section{Appendix for Experiments} 
\subsection{Baseline models}~\label{sec:Baseline_models}
In this section, we introduce six recent instruction-tuned large language models employed to stress-test the Negotiation.
GPT models from OpenAI use a decoder-only transformer framework, and GPT-4 \cite{DBLP:journals/corr/abs-2303-08774}, ChatGPT \citep{openai2022chatgpt} are proprietary models tested by calling the model API. 
Claude-v1.3~\cite{IntroducingClaude1} and Claude-v2.1~\cite{IntroducingClaude2} are closed-source LLMs developed by Anthropic. These two models can be accessible through a chat interface and API, and they demonstrate a strong performance in a lot of NLP tasks. Llama-2 Chat 13B~\cite{DBLP:journals/corr/abs-2307-09288}, and Llama-2 Chat 70B~\cite{DBLP:journals/corr/abs-2307-09288} are a language model fine-tuned for engaging in dialogues that follow user inputs.

\subsection{Hyperparameter} \label{sec:hyparameter}
We use default hyperparameters for all the large language models mentioned in this paper. 
For ChatGPT (\texttt{gpt-3.5-turbo-0613})  and GPT-4 (\texttt{gpt-4-0613}), the default parameters\footnote{https://platform.openai.com/docs/api-reference/chat/create} are \texttt{temperature}=1 and \texttt{top\_p}=1.
For LLaMa2-Chat(13B) and LLaMa2-Chat(70B) models, we follow the default setting where \texttt{temperature}=0.5, \texttt{top\_p}=0.9.
For Claude-v1.3 and Claude-v2.1, the default parameters are \texttt{temperature}=0.5, \texttt{top\_p}=1.


\subsection{Few-Shot Experimental setting} \label{sec:Few-Shot-experimental-setting}
Following~\citet{DBLP:conf/nips/BrownMRSKDNSSAA20}, we use a few-shot prompt to instruct all models:
\begin{center}
\resizebox{.6\linewidth}{!}{
\begin{tabular}{l}
  \textbf{\texttt{ <TASK-PROMPT> }} \\
  \textbf{\texttt{ \textcolor{headcolor}{<EX$_1$-INP>}\textcolor{tailcolor}{<EX$_1$-OUT>} }}
  \\
  ~~\ldots \\
  \textbf{\texttt{ \textcolor{headcolor}{<EX$_{N-1}$-INP>}\textcolor{tailcolor}{<EX$_{N-1}$-OUT>} }}\\
  \textbf{\texttt{ \textcolor{headcolor}{<EX$_N$-INP>} }}
\end{tabular}
}
\end{center}
where \textbf{\texttt{<TASK-PROMPT>}} is a task instruction that explains the background of the negotiation scenario and \textbf{\texttt{\textcolor{headcolor}{<EX$_1$-INP>}\textcolor{tailcolor}{<EX$_1$-OUT>}}} are human-authored examples that try to cover all subclass labels to help LLMs understand the subclass labels.
Finally, we provide the N$_{\text{th}}$ input as \textbf{\texttt{\textcolor{headcolor}{<EX$_N$-INP>}}} and ask all models to generate the corresponding answer as \textbf{\texttt{\textcolor{tailcolor}{<EX$_N$-OUT>}}}.
In this paper, we set $N$ = 4 for the desire and belief states, and $N$ = 7 for the intention state.

\subsection{Appendix for Prompt Template} \label{sec:Prompt_Template}
In this section, we introduce all prompt templates and these templates are presented in Tables~\ref{tab:baseline_prompt_template}, ~\ref{tab:cot_prompt_template}, ~\ref{tab:Fewshot_desire}, ~\ref{tab:Fewshot_belief}, ~\ref{tab:Fewshot_intention},~\ref{tab:ranking_template}, ~\ref{tab:individual_template_1}, 
~\ref{tab:individual_template_2}, ~\ref{tab:strategy_prediction_template}, and ~\ref{tab:strategy_prediction_template_1}. 

\paragraph{Main Result Template} Table~\ref{tab:baseline_prompt_template} introduces the baseline prompt template, a straightforward Q\&A session without requiring detailed and complex explanations. Table~\ref{tab:cot_prompt_template} is the Chain-of-Thought prompt template that facilitates a detailed, step-by-step thought process in the LLM, thoroughly considering each problem aspect before arriving at a solution. Tables~\ref{tab:Fewshot_desire}, ~\ref{tab:Fewshot_belief}, and ~\ref{tab:Fewshot_intention} are presented the few-shot setting prompt template. Each prompt template appends some exemplars for LLMs to learn the new tasks.

\paragraph{Question Format Template} The Ranking Question Format in Table~\ref{tab:ranking_template} asks the LLM to prioritize or rank items, which is useful for tasks needing decision-making based on preference or importance. Table~\ref{tab:individual_template_1} and~\ref{tab:individual_template_2}, the individual question format prompt templates, break down desire and belief questions into separate and focused questions.

\paragraph{Strategy Prediction Template} Tables~\ref{tab:strategy_prediction_template} and~\ref{tab:strategy_prediction_template_1} present the prompt template for the strategy prediction task; the former provides a baseline format for predicting strategies from negotiation conversations, while the latter enriches this task by incorporating the desires and beliefs of the agents involved, offering a deeper contextual understanding and offer signals from desire and belief states for strategy classification.

\subsection{Related Works for Large language Models} 
Recent studies have extensively and comprehensively evaluated instruction following LLMs'~\cite{DBLP:journals/corr/abs-2303-08774, openai2022chatgpt, alpaca, DBLP:conf/emnlp/JiangCCW23} performance on numerous tasks, revealing its superior performance in zero-shot scenarios compared to other models~\cite{
DBLP:journals/corr/abs-2303-12712,
DBLP:journals/corr/abs-2302-04023,
DBLP:conf/eacl/ChanCWJFLS24,
DBLP:conf/emnlp/ChengQCFWCRGZSZ23,
DBLP:conf/acl/0001FLS0XWBLJCS24,
DBLP:conf/coling/JiayangQC0SZ24}. 
However, there are certain obstacles that persist unaddressed, such as the inability to perform complex mathematical reasoning~\cite{DBLP:journals/corr/abs-2301-13867}, 
theory of mind reasoning~\cite{lin2024constrainedreasoningchainsenhancing}, analogies reasoning~\cite{DBLP:conf/emnlp/ChengQCFWCRGZSZ23}, text-to-table generation~\cite{DBLP:journals/corr/abs-2404-14215}, 
fact validation~\cite{DBLP:journals/corr/abs-2310-07521}, complex game setting~\cite{DBLP:journals/corr/abs-2408-02559}, associated ethical implications, and privacy concerns~\cite{DBLP:journals/corr/abs-2310-10383,DBLP:journals/corr/abs-2212-09292,DBLP:conf/acl/0003GLFH0CYYS24,DBLP:journals/corr/abs-2302-00539,DBLP:journals/corr/abs-2405-07667}.
Therefore, it is critical to discuss whether large language models possess the capacity of the theory of mind as humans do. 
In this paper, we test several instruction-following LLMs in our NegotiationToM benchmark. The zero-shot performance of large language models, which relies on the sophisticated design of templates, has shown variance across various tasks~\cite{DBLP:conf/naacl/MaZGTLZH22,
DBLP:conf/ijcnlp/ChanLCCSWS23,
DBLP:conf/acl/ChanLCLSWS23}.
To obtain replicable and representative results, we follow~\citet{DBLP:conf/iclr/RobinsonW23} to formulate the task as a multiple-choice question-answering problem in this work.

\section{Appendix for Experimental Result} \label{sec:appendix_result}

\subsection{Appendix for Intention Subclass Result} \label{sec:Appendix_for_Intention_Result}
\begin{figure*}[t]
    \centering
    \includegraphics[width=1\textwidth]{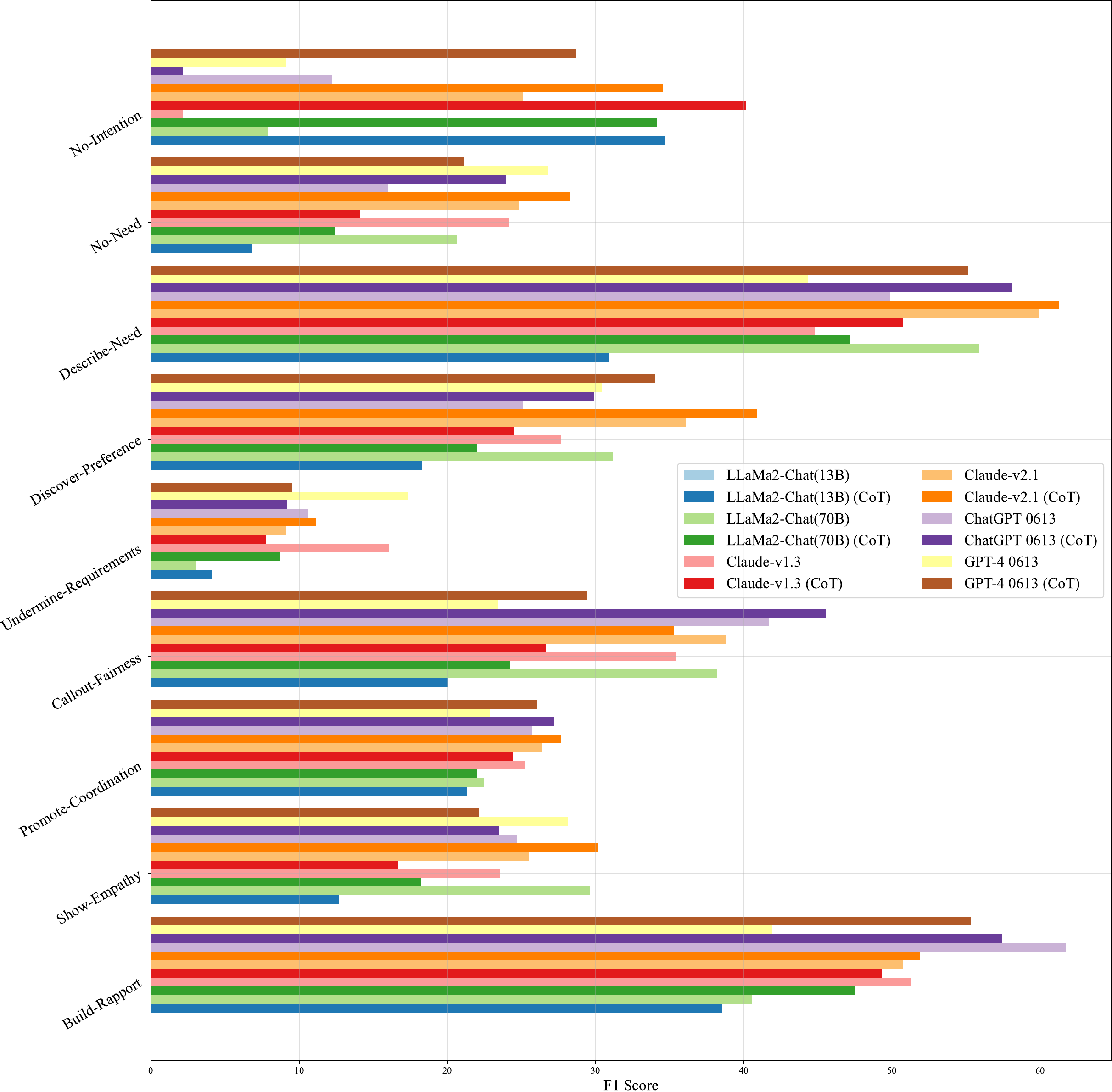}
    \caption{The label-wise intention dimension results of large language models in NegotiationToM.}
    \label{fig:Overall_Intention_Results_labelwise}
\end{figure*}

In this section, Table~\ref{fig:Overall_Intention_Results_labelwise} and Figure~\ref{tab:label-wise_intention} provide more detailed performance of the intention subclass level. All models receive a good performance on the Build-Rapport and Describe-Need subclass while poor at the No-Intention, No-Need, and Undermine-Requirements subclass. 

\subsection{Appendix for Human Performance}\label{sec:Appendix_for_huamn_performance}
To measure the specific performance gap between humans and the state-of-the-art machine on the NegotiationToM, we employ three graduate students in computer science to complete the human evaluation task.
Each instance is shown to three workers who were prompted to choose the more plausible choice for questions regarding desire, belief, and intention. The questions and instructions are the same as the annotation template and are displayed in Appendix~\ref{data_annotation_template}. We compute the majority vote on the labels assigned, which leads to a performance of 91.14\% on both the desire and belief, while the intention only obtained an 83.75\% micro F1 score and 84.65\% macro F1 score. The \textit{"All"} score received 43.78\%, and conversation consistency acquired 75.44\% in both the desire and belief states.

\begin{table*}
\small
\centering
\scalebox{0.9}{\begin{tabular}{|c|c|} \hline 

\multicolumn{1}{|c|}{Models} & Example\\ \hline 
ChatGPT& \parbox[c]{14cm}{
\textbf{Instruction:} You are provided with historical dialogue from the CaSiNO dataset. Finish Next dialogue as appeared in the dataset. Next dialogue must exactly match the instance in the dataset.\newline
\textbf{agent 1:} Hello! Let's work together on a deal for these packages, shall we? What are you most interested in? \newline
\textbf{agent 2:} Hey! I'd like some more firewood to keep my doggo warm. What do you need? \newline
\textbf{agent 1:} I need firewood as well. We have a large group consisting of mostly senior citizens, including my grandma, so we'd like the firewood to keep everyone warm. \newline
\textbf{agent 2:} I see. What are you least interested in? \newline
\hl{\textbf{agent 1:} We are least interested in the hiking equipment. Our group is not very active and prefers to stay indoors.}\newline
}\\\hline
GPT-4& \parbox[c]{14cm}{
\textbf{Instruction:} You are provided with historical dialogue from the CaSiNO dataset. Finish Next dialogue as appeared in the dataset. Next dialogue must exactly match the instance in the dataset.\newline
\textbf{agent 1:} Hello! Let's work together on a deal for these packages, shall we? What are you most interested in? \newline
\textbf{agent 2:} Hey! I'd like some more firewood to keep my doggo warm. What do you need? \newline
\textbf{agent 1:} I need firewood as well. We have a large group consisting of mostly senior citizens, including my grandma, so we'd like the firewood to keep everyone warm. \newline
\textbf{agent 2:} I see. What are you least interested in? \newline
\hl{\textbf{agent 1:} I'm least interested in the food supplies. We have plenty of those.}\newline
}\\\hline

Ground-Truth& \parbox[c]{14cm}{
\textbf{Instruction:} You are provided with historical dialogue from the CaSiNO dataset. Finish Next dialogue as appeared in the dataset. Next dialogue must exactly match the instance in the dataset.\newline
\textbf{agent 1:} Hello! Let's work together on a deal for these packages, shall we? What are you most interested in? \newline
\textbf{agent 2:} Hey! I'd like some more firewood to keep my doggo warm. What do you need? \newline
\textbf{agent 1:} I need firewood as well. We have a large group consisting of mostly senior citizens, including my grandma, so we'd like the firewood to keep everyone warm. \newline
\textbf{agent 2:} I see. What are you least interested in? \newline
\textbf{agent 1: We can make do without extra water. Can we trade two waters for an extra firewood package and an extra food package?.}\newline
}\\\hline

\end{tabular}}
\caption{Prompting template and outcome for verification of potential contamination in CaSiNo dataset. The highlighted sentences are generated by the LLMs.}
\label{tab:verify_casino}
\end{table*}

\begin{table*}[!t]
\small
\centering
\setlength\tabcolsep{4pt}
\begin{tabular}{l|c|c|c|c|c|c|c|c|c}
\toprule
Models & B-R & S-E & P-C & C-F & U-R & D-P & D-N & N-N & N-I\\
\midrule
LLaMa2-Chat(13B)          &24.34 & 15.86 & 21.58 & 32.28 & 2.85 & 26.35 & 33.01 & 8.68 & 13.41              \\
LLaMa2-Chat(13B) (CoT)    &38.54 & 12.66 & 21.33 & 20.02 & 4.09 & 18.26 & 30.89 & 6.84 & \underline{34.64}               \\
LLaMa2-Chat(70B)          &40.56 & \underline{29.60} & 22.44 & 38.19 & 2.99 & 31.19 & 55.89 & 20.62 & 7.85               \\
LLaMa2-Chat(70B) (CoT)    &47.46 & 18.22 & 22.00 & 24.26 & 8.70 & 21.98 & 47.18 & 12.40 & 34.14               \\

\midrule
Claude-v1.3                 &51.27 & 23.55 & 25.26 & 35.44 & \underline{16.06} & 27.65 & 44.79 & 24.13 & 2.14                \\
Claude-v1.3 (CoT)          &49.31 & 16.66 & 24.44 & 26.63 & 7.73 & 24.48 & 50.74 & 14.10 & \textbf{40.16}                \\
Claude-v2.1                 &50.73 & 25.53 & 26.42 & 38.76 & 9.12 & \underline{36.11} & \underline{59.93} & 24.80 & 25.08   \\
Claude-v2.1 (CoT)          &51.86 & \textbf{30.15} & \textbf{27.68} & 35.26 & 11.11 & \textbf{40.91} & \textbf{61.24} & \textbf{28.26} & 34.57     \\
\midrule
ChatGPT 0613               &\textbf{61.72} & 24.69 & 25.73 & \underline{41.71} & 10.61 & 25.07 & 49.87 & 15.99 & 12.20              \\
ChatGPT 0613 (CoT)         &\underline{57.46} & 23.47 & \underline{27.22} & \textbf{45.53} & 9.20 & 29.92 & 58.13 & 23.97 & 2.16            \\
GPT-4 0613                 &41.92 & 28.16 & 22.89 & 23.45 & \textbf{17.32} & 30.42 & 44.30 & \underline{26.77} & 9.12     \\
GPT-4 0613 (CoT)           &55.35 & 22.11 & 26.04 & 29.43 & 9.50 & 34.03 & 55.15 & 21.09 & 28.64             \\
\midrule
\end{tabular}
\caption{Label-wise results of all models for the intention dimension. \textit{w B.D.} indicate with the input with desire and belief. \textit{B-R,  S-E, P-C, C-F, U-R, D-P, D-N, N-N, N-I} stands for Build-Rapport, Show-Empathy, Promote-Coordination, Callout-Fairness, Undermine-Requirements, Discover-Preference, Describe-Need, No-Need, No-Intention. The best results are \textbf{bold-faced}, and the second-best ones are \underline{underlined}.
}
\label{tab:label-wise_intention}
\end{table*}

\begin{table*}
\small
\centering
\scalebox{0.9}{\begin{tabular}{|c|c|} \hline 

\multicolumn{1}{|c|}{Dimension} & Example\\ \hline 

Belief& \parbox[c]{14cm}{Background: Here is a negotiation conversation for a camping trip. There are two agents who own some basic supplies and negotiate with each other to split the additional food packages, water bottles, and firewood to make their camping trip even better. Each of these items will be of either High, Medium or Low priority for these two agents. Each of the additional items only has an available quantity of 3. Please answer the following three questions using "A", "B", "C", "D" without any explanation.\newline Dialogue History:\newline agent 1: Hello! Let's work together on a deal for these packages, shall we? What are you most interested in?\newline agent 2: Hey! I'd like some more firewood to keep my doggo warm. What do you need?\newline Question1: Based on the dialogue, what is the high preference for items Agent 1 thinks Agent 2 is?\newline A.Not given\newline B.Water\newline C.Food\newline D.Firewood\newline Question2: Based on the dialogue, what is the medium preference for items Agent 1 thinks Agent 2 is?\newline A.Not given\newline B.Water\newline C.Food\newline D.Firewood\newline Question3: Based on the dialogue, what is the low preference for items Agent 1 thinks Agent 2 is?\newline A.Not given\newline B.Water\newline C.Food\newline D.Firewood\newline Answer:\newline
}\\\hline
 Desire&\parbox[c]{14cm}{Background: Here is a negotiation conversation for a camping trip. There are two agents who own some basic supplies and negotiate with each other to split the additional food packages, water bottles, and firewood to make their camping trip even better. Each of these items will be of either High, Medium or Low priority for these two agents. Each of the additional items only has an available quantity of 3. Please answer the following three questions using "A", "B", "C", "D" without any explanation.\newline Dialogue History:\newline agent 1: Hello! Let's work together on a deal for these packages, shall we? What are you most interested in?\newline agent 2: Hey! I'd like some more firewood to keep my doggo warm. What do you need?\newline Question1: What is agent 1's high preference for items based on the dialogue history?\newline A.Not given\newline B.Water\newline C.Food\newline D.Firewood\newline Question2: What is agent 1's medium preference for items based on the dialogue history?\newline A.Not given\newline B.Water\newline C.Food\newline D.Firewood\newline Question3: What is agent 1's low preference for items based on the dialogue history?\newline A.Not given\newline B.Water\newline C.Food\newline D.Firewood\newline Answer:\newline
} \\\hline 
 Intention&\parbox[c]{14cm}{Background: Here is a negotiation conversation for a camping trip. There are two agents who own some basic supplies and negotiate with each other to split the additional food packages, water bottles, and firewood to make their camping trip even better. Each of these items will be of either High, Medium or Low priority for these two agents. Each of the additional items only has an available quantity of 3.\newline Dialogue History:\newline agent 1: Hello! Let's work together on a deal for these packages, shall we? What are you most interested in?\newline agent 2: Hey! I'd like some more firewood to keep my doggo warm. What do you need?\newline Question: What are the plausible intentions of Agent 1 expressed in 'Hello! Let's work together on a deal for these packages, shall we? What are you most interested in?' Based on the dialogue history, select one or more intentions (i.e.,"A", "B", "C",...,"I") from the following choices without any explanation.\newline A.Intents to build a rapport with the opponent\newline B.Intents to show empathy with the opponent\newline C.Intents to promote coordination with the opponent\newline D.Intents to callout to fairness\newline E.Intents to undermine the requirements of the opponent\newline F.Intents to discover the preference order of the opponent\newline G.Intents to describe a need for an item\newline H.Intents to point out they do not need an item\newline I.No clear intention in the utterance\newline Answer:\newline
}\\\hline

\end{tabular}}
\caption{Baseline prompt template.}
\label{tab:baseline_prompt_template}
\end{table*}

\begin{table*}
\small
\centering
\scalebox{0.9}{\begin{tabular}{|c|c|} \hline 

\multicolumn{1}{|c|}{Dimension} & Example\\ \hline 

Belief& \parbox[c]{14cm}{
Background: Here is a negotiation conversation for a camping trip. There are two agents who own some basic supplies and negotiate with each other to split the additional food packages, water bottles, and firewood to make their camping trip even better. Each of these items will be of either High, Medium or Low priority for these two agents. Each of the additional items only has an available quantity of 3. Please answer the following three questions using "A", "B", "C", "D". \newline Dialogue History:\newline agent 1: Hello!  Let's work together on a deal for these packages, shall we? What are you most interested in?\newline agent 2: Hey! I'd like some more firewood to keep my doggo warm. What do you need?\newline Question1: Based on the dialogue, what is the high preference for items Agent 1 thinks Agent 2 is?\newline A.Not given\newline B.Water\newline C.Food\newline D.Firewood\newline Question2: Based on the dialogue, what is the medium preference for items Agent 1 thinks Agent 2 is?\newline A.Not given\newline B.Water\newline C.Food\newline D.Firewood\newline Question3: Based on the dialogue, what is the low preference for items Agent 1 thinks Agent 2 is?\newline A.Not given\newline B.Water\newline C.Food\newline D.Firewood\newline Answer: Let’s think step by step.\newline
}\\ \hline
 Desire&\parbox[c]{14cm}{
 Background: Here is a negotiation conversation for a camping trip. There are two agents who own some basic supplies and negotiate with each other to split the additional food packages, water bottles, and firewood to make their camping trip even better. Each of these items will be of either High, Medium or Low priority for these two agents. Each of the additional items only has an available quantity of 3. Please answer the following three questions using "A", "B", "C", "D".\newline Dialogue History:\newline agent 1: Hello! Let's work together on a deal for these packages, shall we? What are you most interested in?\newline agent 2: Hey! I'd like some more firewood to keep my doggo warm. What do you need?\newline Question1: What is agent 1's high preference for items based on the dialogue history?\newline A.Not given\newline B.Water\newline C.Food\newline D.Firewood\newline Question2: What is agent 1's medium preference for items based on the dialogue history?\newline A.Not given\newline B.Water\newline C.Food\newline D.Firewood\newline Question3: What is agent 1's low preference for items based on the dialogue history?\newline A.Not given\newline B.Water\newline C.Food\newline D.Firewood\newline Answer: Let’s think step by step.\newline
 }\\\hline
 Intention&\parbox[c]{14cm}{
 Background: Here is a negotiation conversation for a camping trip. There are two agents who own some basic supplies and negotiate with each other to split the additional food packages, water bottles, and firewood to make their camping trip even better. Each of these items will be of either High, Medium or Low priority for these two agents. Each of the additional items only has an available quantity of 3.\newline Dialogue History:\newline agent 1: Hello! Let's work together on a deal for these packages, shall we? What are you most interested in?\newline agent 2: Hey! I'd like some more firewood to keep my doggo warm. What do you need?\newline Question: What are the plausible intentions of Agent 1 expressed in 'Hello!  Let's work together on a deal for these packages, shall we? What are you most interested in?' Based on the dialogue history, select one or more intentions (i.e.,"A", "B", "C",...,"I") from the following choices.\newline A.Intents to build a rapport with the opponent\newline B.Intents to show empathy with the opponent\newline C.Intents to promote coordination with the opponent\newline D.Intents to callout to fairness\newline E.Intents to undermine the requirements of the opponent\newline F.Intents to discover the preference order of the opponent\newline G.Intents to describe a need for an item\newline H.Intents to point out they do not need an item\newline I.No clear intention in the utterance\newline Answer: Let’s think step by step.\newline
 }\\\hline

\end{tabular}}
\caption{Chain-of-Thought prompt template.}
\label{tab:cot_prompt_template}
\end{table*}

\begin{table*}
\small
\centering
\scalebox{0.9}{\begin{tabular}{|c|c|} \hline 

\multicolumn{1}{|c|}{Dimension} & Example\\ \hline 
Desire& \parbox[c]{14cm}{Background: Here is a negotiation conversation for a camping trip. There are two agents who own some basic supplies and negotiate with each other to split the additional food packages, water bottles, and firewood to make their camping trip even better. Each of these items will be of either High, Medium or Low priority for these two agents. Each of the additional items only has an available quantity of 3. Please answer the following three questions using "A", "B", "C", "D" without any explanation.\newline
Dialogue History:\newline
agent 1: Hello! Which item do you need the most?\newline
agent 2: I would love to have the Firewood the most. \newline
Question1: What is agent 1's high preference for items based on the dialogue history?\newline
A.Not given\newline
B.Water\newline
C.Food\newline
D.Firewood\newline
Question2: What is agent 1's medium preference for items based on the dialogue history?\newline
A.Not given\newline
B.Water\newline
C.Food\newline
D.Firewood\newline
Question3: What is agent 1's low preference for items based on the dialogue history?\newline
A.Not given\newline
B.Water\newline
C.Food\newline
D.Firewood\newline
Answer: A,A,A

Dialogue History:\newline
agent 1: Hello! Which item do you need the most?\newline
agent 2: I would love to have the Firewood the most. \newline
agent 1: Unfortunately I need firewood the most too. How about I take 2 firewood, 2 food, and 1 water?\newline
agent 2: I feel that I am not getting a fair deal.\newline 
Question1: What is agent 1's high preference for items based on the dialogue history?\newline
A.Not given\newline
B.Water\newline
C.Food\newline
D.Firewood\newline
Question2: What is agent 1's medium preference for items based on the dialogue history?\newline
A.Not given\newline
B.Water\newline
C.Food\newline
D.Firewood\newline
Question3: What is agent 1's low preference for items based on the dialogue history?\newline
A.Not given\newline
B.Water\newline
C.Food\newline
D.Firewood\newline
Answer: D,C,B\newline
...\newline
Dialogue History:\newline
agent 1: Hello! Let's work together on a deal for these packages, shall we? What are you most interested in?\newline
agent 2: Hey! I'd like some more firewood to keep my doggo warm. What do you need?\newline
Question1: What is agent 1's high preference for items based on the dialogue history?\newline
A.Not given\newline
B.Water\newline
C.Food\newline
D.Firewood\newline
Question2: What is agent 1's medium preference for items based on the dialogue history?\newline
A.Not given\newline
B.Water\newline
C.Food\newline
D.Firewood\newline
Question3: What is agent 1's low preference for items based on the dialogue history?\newline
A.Not given\newline
B.Water\newline
C.Food\newline
D.Firewood\newline
Answer:\newline
}\\\hline

\end{tabular}}
\caption{Few-shot prompt template for desire state.}
\label{tab:Fewshot_desire}
\end{table*}

\begin{table*}
\small
\centering
\scalebox{0.9}{\begin{tabular}{|c|c|} \hline 

\multicolumn{1}{|c|}{Dimension} & Example\\ \hline 
Desire& \parbox[c]{14cm}{Background: Here is a negotiation conversation for a camping trip. There are two agents who own some basic supplies and negotiate with each other to split the additional food packages, water bottles, and firewood to make their camping trip even better. Each of these items will be of either High, Medium or Low priority for these two agents. Each of the additional items only has an available quantity of 3. Please answer the following three questions using "A", "B", "C", "D" without any explanation.\newline
Dialogue History:\newline
agent 1: Hello! Which item do you need the most?\newline
agent 2: I would love to have the Firewood the most.\newline
Question1: Based on the dialogue, what is the high preference for items Agent 1 thinks Agent 2 is?\newline
A.Not given\newline
B.Water\newline
C.Food\newline
D.Firewood\newline
Question2: Based on the dialogue, what is the medium preference for items Agent 1 thinks Agent 2 is?\newline
A.Not given\newline
B.Water\newline
C.Food\newline
D.Firewood\newline
Question3: Based on the dialogue, what is the low preference for items Agent 1 thinks Agent 2 is?\newline
A.Not given\newline
B.Water\newline
C.Food\newline
D.Firewood\newline
Answer: D,A,A\newline
Dialogue History:\newline
agent 1: Hello! Which item do you need the most?\newline
agent 2: I would love to have the Firewood the most.\newline
agent 1: Unfortunately I need firewood the most too. How about I take 2 firewood, 2 food, and 1 water?\newline
agent 2: I feel that I am not getting a fair deal. \newline
agent 1: Then what do you think is fair?\newline
agent 2: I think that I should get 2 firewood, 1 food and 2 water\newline
Question1: Based on the dialogue, what is the high preference for items Agent 1 thinks Agent 2 is?\newline
A.Not given\newline
B.Water\newline
C.Food\newline
D.Firewood\newline
Question2: Based on the dialogue, what is the medium preference for items Agent 1 thinks Agent 2 is?\newline
A.Not given\newline
B.Water\newline
C.Food\newline
D.Firewood\newline
Question3: Based on the dialogue, what is the low preference for items Agent 1 thinks Agent 2 is?\newline
A.Not given\newline
B.Water\newline
C.Food\newline
D.Firewood\newline
Answer: D,B,C\newline
...\newline
Dialogue History:\newline
agent 1: Hello! Let's work together on a deal for these packages, shall we? What are you most interested in?\newline
agent 2: Hey! I'd like some more firewood to keep my doggo warm. What do you need?\newline
Question1: Based on the dialogue, what is the high preference for items Agent 1 thinks Agent 2 is?\newline
A.Not given\newline
B.Water\newline
C.Food\newline
D.Firewood\newline
Question2: Based on the dialogue, what is the medium preference for items Agent 1 thinks Agent 2 is?\newline
A.Not given\newline
B.Water\newline
C.Food\newline
D.Firewood\newline
Question3: Based on the dialogue, what is the low preference for items Agent 1 thinks Agent 2 is?\newline
A.Not given\newline
B.Water\newline
C.Food\newline
D.Firewood\newline
Answer:\newline
}\\\hline

\end{tabular}}
\caption{Few-shot prompt template for belief state.}
\label{tab:Fewshot_belief}
\end{table*}

\begin{table*}
\small
\centering
\scalebox{0.9}{\begin{tabular}{|c|c|} \hline 

\multicolumn{1}{|c|}{Dimension} & Example\\ \hline 
Desire& \parbox[c]{14cm}{Background: Here is a negotiation conversation for a camping trip. There are two agents who own some basic supplies and negotiate with each other to split the additional food packages, water bottles, and firewood to make their camping trip even better. Each of these items will be of either High, Medium or Low priority for these two agents. Each of the additional items only has an available quantity of 3.\newline
Dialogue History:\newline
agent 2: Hi! I'm super excited to go camping with my family as a great way to vacation due to Covid19. My kid is so restless from being cooped up in the house all the time. Are you planning on going camping too?\newline
agent 1: I am! It is the perfect way to get away and still manage to social distance! I am worried about having enough water though, are you short on any supplies?\newline
agent 2: I think I'm good. I'm not 100\% sure. My husband likes to do adventures on the fly. He got these water filter straw thingies from Amazon and said that if we run out of the water I packed, that we can drink the water in the lake but I don't really trust the straws.\newline
agent 1: Sounds like you need water too. How about you take 2 water, 1 food, and 1 firewood?\newline
Question: What are the plausible intentions of Agent 2 expressed in 'I think I'm good. I'm not 100\% sure. My husband likes to do adventures on the fly. He got these water filter straw thingies from Amazon and said that if we run out of the water I packed, that we can drink the water in the lake but I don't really trust the straws.' Based on the dialogue history, select one or more intentions (i.e.,"A", "B", "C",...,"I") from the following choices without any explanation.\newline
A.Intents to build a rapport with the opponent\newline
B.Intents to show empathy with the opponent\newline
C.Intents to promote coordination with the opponent\newline
D.Intents to callout to fairness\newline
E.Intents to undermine the requirements of the opponent\newline
F.Intents to discover the preference order of the opponent\newline
G.Intents to describe a need for an item\newline
H.Intents to point out they do not need an item\newline
I.No clear intention in the utterance\newline
Answer: A,H\newline
Dialogue History:\newline
agent 2: Looking forward to this camping trip!  I am hoping we can find something amicable with these additional resources.\newline
agent 1: I'm excited too. Things have been stressful lately. What are some things that you value most?\newline
Question: What are the plausible intentions of Agent 2 expressed in 'Looking forward to this camping trip!  I am hoping we can find something amicable with these additional resources.' Based on the dialogue history, select one or more intentions (i.e.,"A", "B", "C",...,"I") from the following choices without any explanation.\newline
A.Intents to build a rapport with the opponent\newline
B.Intents to show empathy with the opponent\newline
C.Intents to promote coordination with the opponent\newline
D.Intents to callout to fairness\newline
E.Intents to undermine the requirements of the opponent\newline
F.Intents to discover the preference order of the opponent\newline
G.Intents to describe a need for an item\newline
H.Intents to point out they do not need an item\newline
I.No clear intention in the utterance\newline
Answer: A,C\newline
...\newline
Dialogue History:\newline
agent 1: Hello! Let's work together on a deal for these packages, shall we? What are you most interested in?\newline
agent 2: Hey! I'd like some more firewood to keep my doggo warm. What do you need?\newline
Question: What are the plausible intentions of Agent 1 expressed in 'Hello! Let's work together on a deal for these packages, shall we? What are you most interested in?' Based on the dialogue history, select one or more intentions (i.e.,"A", "B", "C",...,"I") from the following choices without any explanation.\newline
A.Intents to build a rapport with the opponent\newline
B.Intents to show empathy with the opponent\newline
C.Intents to promote coordination with the opponent\newline
D.Intents to callout to fairness\newline
E.Intents to undermine the requirements of the opponent\newline
F.Intents to discover the preference order of the opponent\newline
G.Intents to describe a need for an item\newline
H.Intents to point out they do not need an item\newline
I.No clear intention in the utterance\newline
Answer:\newline
}\\\hline

\end{tabular}}
\caption{Few-shot prompt template for intention state.}
\label{tab:Fewshot_intention}
\end{table*}

\begin{table*}
\small
\centering
\scalebox{0.9}{\begin{tabular}{|c|c|} \hline 
\multicolumn{1}{|c|}{Dimension} & Example\\ \hline 

Belief& \parbox[c]{14cm}{
Background: Here is a negotiation conversation for a camping trip. There are two agents who own some basic supplies and negotiate with each other to split the additional food packages, water bottles, and firewood to make their camping trip even better. Each of these items will be of either High, Medium or Low priority for these two agents. Each of the additional items only has an available quantity of 3. Please answer the following question without any explanation by ordering the agent prefernece order using A represent "Not given", B represent "Water", C represent "Food", and D represent "Firewood". For exmple A,B,C means that the high preferece item is "Not given", the "Water" is medium preference item, and the "Food" is low preference item.\newline
Dialogue History:\newline
agent 1: Hello! Let's work together on a deal for these packages, shall we? What are you most interested in?\newline
agent 2: Hey! I'd like some more firewood to keep my doggo warm. What do you need?\newline
agent 1: I need firewood as well. We have a large group consisting of mostly senior citizens, including my grandma, so we'd like the firewood to keep everyone warm.\newline
agent 2: I see. What are you least interested in?\newline
Question: Based on the dialogue, what is the high preference for items Agent 1 thinks Agent 2 is?\newline
1. A,A,A\newline
2. A,A,B\newline
3. A,A,C\newline
4. A,A,D\newline
5. A,B,A\newline
6. A,B,C\newline
7. A,B,D\newline
8. A,C,A\newline
9. A,C,B\newline
10. A,C,D\newline
11. A,D,A\newline
12. A,D,B\newline
13. A,D,C\newline
14. B,A,A\newline
15. B,C,D\newline
16. B,D,C\newline
17. C,A,A\newline
18. C,B,D\newline
19. C,D,B\newline
20. D,A,A\newline
21. D,B,C\newline
22. D,C,B\newline
Answer:\newline
}\\ \hline
 Desire&\parbox[c]{14cm}{
Background: Here is a negotiation conversation for a camping trip. There are two agents who own some basic supplies and negotiate with each other to split the additional food packages, water bottles, and firewood to make their camping trip even better. Each of these items will be of either High, Medium or Low priority for these two agents. Each of the additional items only has an available quantity of 3. Please answer the following question without any explanation by ordering the agent prefernece order using A represent "Not given", B represent "Water", C represent "Food", and D represent "Firewood". For exmple A,B,C means that the high preferece item is "Not given", the "Water" is medium preference item, and the "Food" is low preference item.\newline
Dialogue History:\newline
agent 1: Hello! Let's work together on a deal for these packages, shall we? What are you most interested in?\newline
agent 2: Hey! I'd like some more firewood to keep my doggo warm. What do you need?\newline
agent 1: I need firewood as well. We have a large group consisting of mostly senior citizens, including my grandma, so we'd like the firewood to keep everyone warm.\newline
agent 2: I see. What are you least interested in?\newline
Question: What is agent 1's preference order for items based on the dialogue history?\newline
1. A,A,A\newline
2. A,A,B\newline
3. A,A,C\newline
4. A,A,D\newline
5. A,B,A\newline
6. A,B,C\newline
7. A,B,D\newline
8. A,C,A\newline
9. A,C,B\newline
10. A,C,D\newline
11. A,D,A\newline
12. A,D,B\newline
13. A,D,C\newline
14. B,A,A\newline
15. B,C,D\newline
16. B,D,C\newline
17. C,A,A\newline
18. C,B,D\newline
19. C,D,B\newline
20. D,A,A\newline
21. D,B,C\newline
22. D,C,B\newline
Answer:\newline
 }\\\hline

\end{tabular}}
\caption{Ranking question format prompt template.}
\label{tab:ranking_template}
\end{table*}

\begin{table*}
\small
\centering
\scalebox{0.9}{\begin{tabular}{|c|c|} \hline 

\multicolumn{1}{|c|}{Dimension} & Example\\ \hline 

Belief& \parbox[c]{14cm}{Background: Here is a negotiation conversation for a camping trip. There are two agents who own some basic supplies and negotiate with each other to split the additional food packages, water bottles, and firewood to make their camping trip even better. Each of these items will be of either High, Medium or Low priority for these two agents. Each of the additional items only has an available quantity of 3. Please answer the following three questions using "A", "B", "C", "D" without any explanation.\newline Dialogue History:\newline agent 1: Hello! Let's work together on a deal for these packages, shall we? What are you most interested in?\newline agent 2: Hey! I'd like some more firewood to keep my doggo warm. What do you need?\newline Question: Based on the dialogue, what is the high preference for items Agent 1 thinks Agent 2 is?\newline A.Not given\newline B.Water\newline C.Food\newline D.Firewood\newline Answer:\newline
}\\\hline
Belief& \parbox[c]{14cm}{Background: Here is a negotiation conversation for a camping trip. There are two agents who own some basic supplies and negotiate with each other to split the additional food packages, water bottles, and firewood to make their camping trip even better. Each of these items will be of either High, Medium or Low priority for these two agents. Each of the additional items only has an available quantity of 3. Please answer the following three questions using "A", "B", "C", "D" without any explanation.\newline Dialogue History:\newline agent 1: Hello! Let's work together on a deal for these packages, shall we? What are you most interested in?\newline agent 2: Hey! I'd like some more firewood to keep my doggo warm. What do you need?\newline Question: Based on the dialogue, what is the medium preference for items Agent 1 thinks Agent 2 is?\newline A.Not given\newline B.Water\newline C.Food\newline D.Firewood\newline Answer:\newline
}\\\hline
Belief& \parbox[c]{14cm}{Background: Here is a negotiation conversation for a camping trip. There are two agents who own some basic supplies and negotiate with each other to split the additional food packages, water bottles, and firewood to make their camping trip even better. Each of these items will be of either High, Medium or Low priority for these two agents. Each of the additional items only has an available quantity of 3. Please answer the following three questions using "A", "B", "C", "D" without any explanation.\newline Dialogue History:\newline agent 1: Hello! Let's work together on a deal for these packages, shall we? What are you most interested in?\newline agent 2: Hey! I'd like some more firewood to keep my doggo warm. What do you need?\newline Question: Based on the dialogue, what is the low preference for items Agent 1 thinks Agent 2 is?\newline A.Not given\newline B.Water\newline C.Food\newline D.Firewood\newline Answer:\newline
}\\\hline

\end{tabular}}
\caption{Individual question format prompt template (I).}
\label{tab:individual_template_1}
\end{table*}

\begin{table*}
\small
\centering
\scalebox{0.9}{\begin{tabular}{|c|c|} \hline 

\multicolumn{1}{|c|}{Dimension} & Example\\ \hline 

Desire& \parbox[c]{14cm}{Background: Here is a negotiation conversation for a camping trip. There are two agents who own some basic supplies and negotiate with each other to split the additional food packages, water bottles, and firewood to make their camping trip even better. Each of these items will be of either High, Medium or Low priority for these two agents. Each of the additional items only has an available quantity of 3. Please answer the following three questions using "A", "B", "C", "D" without any explanation.\newline Dialogue History:\newline agent 1: Hello! Let's work together on a deal for these packages, shall we? What are you most interested in?\newline agent 2: Hey! I'd like some more firewood to keep my doggo warm. What do you need?\newline Question: What is agent 1's high preference for items based on the dialogue history?\newline A.Not given\newline B.Water\newline C.Food\newline D.Firewood\newline Answer:\newline
}\\\hline
Desire& \parbox[c]{14cm}{Background: Here is a negotiation conversation for a camping trip. There are two agents who own some basic supplies and negotiate with each other to split the additional food packages, water bottles, and firewood to make their camping trip even better. Each of these items will be of either High, Medium or Low priority for these two agents. Each of the additional items only has an available quantity of 3. Please answer the following three questions using "A", "B", "C", "D" without any explanation.\newline Dialogue History:\newline agent 1: Hello! Let's work together on a deal for these packages, shall we? What are you most interested in?\newline agent 2: Hey! I'd like some more firewood to keep my doggo warm. What do you need?\newline Question: What is agent 1's medium preference for items based on the dialogue history?\newline A.Not given\newline B.Water\newline C.Food\newline D.Firewood\newline Answer:\newline
}\\\hline
Desire& \parbox[c]{14cm}{Background: Here is a negotiation conversation for a camping trip. There are two agents who own some basic supplies and negotiate with each other to split the additional food packages, water bottles, and firewood to make their camping trip even better. Each of these items will be of either High, Medium or Low priority for these two agents. Each of the additional items only has an available quantity of 3. Please answer the following three questions using "A", "B", "C", "D" without any explanation.\newline Dialogue History:\newline agent 1: Hello! Let's work together on a deal for these packages, shall we? What are you most interested in?\newline agent 2: Hey! I'd like some more firewood to keep my doggo warm. What do you need?\newline Question: What is agent 1's low preference for items based on the dialogue history?\newline A.Not given\newline B.Water\newline C.Food\newline D.Firewood\newline Answer:\newline
}\\\hline

\end{tabular}}
\caption{Individual question format prompt template(II).}
\label{tab:individual_template_2}
\end{table*}

\begin{table*}
\small
\centering
\scalebox{0.9}{\begin{tabular}{|c|c|} \hline 

\multicolumn{1}{|c|}{Dimension} & Example\\ \hline 

Desire& \parbox[c]{14cm}{Background: Here is a negotiation conversation for a camping trip. There are two agents who own some basic supplies and negotiate with each other to split the additional food packages, water bottles, and firewood to make their camping trip even better. Each of these items will be of either High, Medium or Low priority for these two agents. Each of the additional items only has an available quantity of 3.\newline 
Dialogue History:\newline
agent 1: Hello! Let’s work together on a deal for these packages, shall we? What are you most interested in?
agent 2: Hey! I’d like some more firewood to keep my doggo warm. What do you need?\newline
Question: What are the plausible strategies of Agent 1 expressed in ‘Hello! Let’s work together on a deal
for these packages, shall we? What are you most interested in?’. Based on the dialogue history, select one or more strategies (i.e.,"A", "B", "C",...,"J") from the following choices and their definition. Please select "A", "B", "C",...,"J" without any explanation.\newline
A.Small-Talk: Participants discussing topics apart from the negotiation, in an attempt to build a rapport with the partner.\newline
B.Empathy: An utterance depicts Empathy when there is evidence of positive acknowledgments or empathetic behavior towards a personal context of the partner.\newline
C.Coordination: is used when a participant promotes coordination among the two partners.\newline
D.Vouch-Fairness: is a callout to fairness for personal benefit, either when acknowledging a fair deal or when the opponent offers a deal that benefits them.\newline
E.Undervalue-Partner: refers to the scenario where a participant undermines the requirements of their opponent.\newline
F.Elicit-Pref: an attempt to discover the preference order of the opponent.\newline
G.Self-Need: refers to arguments for creating a personal need for an item in the negotiation.\newline
H.Other-Need: used when the participants discuss a need for someone else rather than themselves.\newline
I.No-Need: is when a participant points out that they do not need an item based on personal context.\newline
J.Non-strategic: if no strategy is evident, the utterance is labeled as Non-strategic.\newline
Answer:
}\\\hline

\end{tabular}}
\caption{Baseline prompt template for strategy prediction}
\label{tab:strategy_prediction_template}
\end{table*}

\begin{table*}
\small
\centering
\scalebox{0.9}{\begin{tabular}{|c|c|} \hline 

\multicolumn{1}{|c|}{Dimension} & Example\\ \hline 

Desire& \parbox[c]{14cm}{Background: Here is a negotiation conversation for a camping trip. There are two agents who own some basic supplies and negotiate with each other to split the additional food packages, water bottles, and firewood to make their camping trip even better. Each of these items will be of either High, Medium or Low priority for these two agents. Each of the additional items only has an available quantity of 3.\newline 
Dialogue History:\newline
agent 1: Hello! Let’s work together on a deal for these packages, shall we? What are you most interested in?
agent 2: Hey! I’d like some more firewood to keep my doggo warm. What do you need?\newline
Question: What are the plausible strategies of Agent 1 expressed in ’Hello! Let’s work together on a deal for these packages, shall we? What are you most interested in?’. \textbf{Agent 1's preference order of items is Not Given, Not Given, and Not Given. Agent 2's preference order of items is Firewood, Not Given, and Not Given.} Please imagine that you are Agent 1 and infer your strategies expressed in  ‘Hello! Let’s work together on a deal for these packages, shall we? What are you most interested in?’ by using Agent 1's preference order, Agent 2's preference order, and all information expressed in the dialogue history. Select one or more strategies (i.e.,"A", "B", "C",...,"J") from the following choices and their definition.\newline
A.Small-Talk: Participants discussing topics apart from the negotiation, in an attempt to build a rapport with the partner.\newline
B.Empathy: An utterance depicts Empathy when there is evidence of positive acknowledgments or empathetic behavior towards a personal context of the partner.\newline
C.Coordination: is used when a participant promotes coordination among the two partners.\newline
D.Vouch-Fairness: is a callout to fairness for personal benefit, either when acknowledging a fair deal or when the opponent offers a deal that benefits them.\newline
E.Undervalue-Partner: refers to the scenario where a participant undermines the requirements of their opponent.\newline
F.Elicit-Pref: an attempt to discover the preference order of the opponent.\newline
G.Self-Need: refers to arguments for creating a personal need for an item in the negotiation.\newline
H.Other-Need: used when the participants discuss a need for someone else rather than themselves.\newline
I.No-Need: is when a participant points out that they do not need an item based on personal context.\newline
J.Non-strategic: if no strategy is evident, the utterance is labeled as Non-strategic.\newline
Answer:
}\\\hline

\end{tabular}}
\caption{Prompt template with desire and belief for strategy prediction}
\label{tab:strategy_prediction_template_1}
\end{table*}

\begin{table*}[!] 
\small
\centering
\setlength\tabcolsep{4pt}
\begin{tabular}{lcccccc}
\toprule
Dataset  & Total\#Questions    & Avg.\#Questions per Context   & Avg.\#Turns(Full)  & Avg.TurnLength \\ 
\midrule  
ToMi    & 6K            & 6.0         & 4.9       & 4.7 \\ 
FanToM  & 10K           & 12.9        & 24.5      & 21.9 \\ 
NegotiationToM & 13K    & 7.0         & 6.0       & 42.2 \\ 
\bottomrule
\end{tabular}
\caption{Statistics of our benchmark, FanToM\cite{DBLP:conf/emnlp/0002SZ0K0S23}, and ToMi \cite{le-etal-2019-revisiting}.}
\label{tab:dataset_stats}
\end{table*}

\begin{table*}[th]
\centering
\scalebox{0.7}{
\begin{tabular}{p{4cm}|p{15cm}}
\hline
\multicolumn{1}{c}{\textbf{Strategies}} & \multicolumn{1}{c}{\textbf{Definition}}\\ 
\hline

Small-Talk   & Participants engage in small talk while discussing topics apart from the negotiation, in an attempt to build a rapport with the partner.\\\hline
Empathy      & An utterance depicts Empathy when there is evidence of positive acknowledgments or empathetic behavior towards a personal context of the partner, for instance, towards a medical emergency.\\\hline
Coordination &  is used when a participant promotes coordination among the two partners.\\ \hline
No-Need      & is when a participant points out that they do not need an item based on personal context.\\\hline

Elicit-Pref         & an attempt to discover the preference order of the opponent. \\ \hline
Undervalue-Partner  & refers to the scenario where a participant undermines the requirements of their opponent. \\\hline
Vouch-Fairness      & is a callout to fairness for personal benefit, either when acknowledging a fair deal or when the opponent offers a deal that benefits them.\\ \hline
Self-Need           & refers to arguments for creating a personal need for an item in the negotiation.\\\hline
Other-Need          & is similar to Self-Need but is used when the participants discuss a need for someone else rather than themselves. \\ \hline
Non-strategic       & If no strategy is evident, the utterance is labeled as Non-strategic. \\ \hline
\end{tabular}
}
\caption{\label{tab:strategies_defin}Utterance-level strategy definition as refer to~\citet{DBLP:conf/naacl/ChawlaRCLMG21}.}
\end{table*}

\begin{table*}[th]
\centering
\scalebox{0.7}{
\begin{tabular}{p{4cm}|p{12cm}|p{0.9cm} p{0.9cm}}
\hline
\multicolumn{1}{c}{\textbf{Strategies}} & \multicolumn{1}{c}{\textbf{Example}} & \textbf{Count} & \textbf{$\alpha$}\\ 
\hline
Small-Talk & Hello, how are you today? & $1054$ & $0.81$\\
Empathy & Oh I wouldn't want for you to freeze & $254$ & $0.42$\\
Coordination &  Let's try to make a deal that benefits us both! & $579$ & $0.42$\\ \hline
No-Need & We have plenty of water to spare. & $196$ & $0.77$\\
Elicit-Pref & What supplies do you prefer to take the most of? & $377$ & $0.77$\\ \hline
Undervalue-Partner & Do you have help carrying all that extra firewood? Could be heavy? & $131$ & $0.72$ \\
Vouch-Fairness & That would leave me with no water. & $439$ & $0.62$\\ \hline
Self-Need & I can't take cold and would badly need to have more firewood. & $964$ & $0.75$\\
Other-Need & we got kids on this trip, they need food too. & $409$ & $0.89$\\ \hline
Non-strategic & Hello, I need supplies for the trip! & $1455$ & - \\ \hline
\end{tabular}
}
\caption{\label{tab:ann-stats}Utterance-level negotiation strategy annotations, refer to~\citet{DBLP:conf/naacl/ChawlaRCLMG21}. $\alpha$ refers to Krippendorff's alpha among $3$ annotators on a subset of $10$ dialogues ($\sim120$ utterances). An utterance can have multiple labels.}
\end{table*}

\begin{table*}[ht]
\centering
\scalebox{0.8}{
\begin{tabular}{l|m{5cm}|m{5cm}|m{5cm}}
\hline
\multirow{2}{*}{\textbf{Category}} & \multicolumn{3}{c}{\textbf{Item type}} \\
& \multicolumn{1}{c}{\textbf{Food}} & \multicolumn{1}{c}{\textbf{Water}} & \multicolumn{1}{c}{\textbf{Firewood}} \\ \hline

\textbf{Personal Care} & because I'm normally eat more because of my big size & I have to take a lot of medicine so hydration is very important &  I have arthritis and being sure I am warm is important for my comfort. \\ \hline
\textbf{Recreational} & Need many snacks throughout the day for energy to hike & I am a very active camper. I like to hike when I camp and I once ran out of water during a strenuous hike. &  I like having campfires so I need all the firewood.\\ \hline
\textbf{Group Needs} & I have two teenage boys who require a lot of food, especially when expending so much energy with all the activities of camping. &  I need more water because I have more people to keep hydrated and do not have enough. & I need more firewood due to having several people join on the trip and needing a bigger fire overall. \\ \hline
\textbf{Emergency} & Some could have been damaged during the trip. I would need more. & our car overheated we had to use the water &  It may get cold and firewood can be hard to come by at certain campsites. 
 \\ \hline
\end{tabular}}
\caption{Example arguments that the participants come up for their individual requirements during the preparation phase. The categories defined are not exhaustive.}
\label{tab:sample-reasons}
\end{table*}

\end{document}